\documentclass[lettersize,journal]{IEEEtran}
\usepackage{bm}
\usepackage{amsfonts}
\usepackage{amsmath}
\usepackage{cite}
\usepackage{algorithm}
\usepackage{algpseudocode}
\usepackage{amssymb}
\usepackage{graphicx}
\usepackage{pifont}
\usepackage{amsthm,amssymb}
\usepackage{mathrsfs}
\usepackage{setspace}
\usepackage{color}
\usepackage{longtable}
\usepackage{booktabs}
\usepackage{makecell}
\usepackage[normalem]{ulem}

\usepackage{CJKutf8}
\definecolor{bluedefrance}{rgb}{0.19, 0.55, 0.91}
\definecolor{awesome}{rgb}{1.0, 0.13, 0.32}
\definecolor{forestgreen}{rgb}{0.13, 0.55, 0.13}

\newcommand{\bmc}[1]{\boldsymbol{\mathcal{#1}}}

\begin{document}
\title{Direction-Constrained Control for Efficient Physical Human-Robot Interaction under Hierarchical Tasks}
\author{Mengxin Xu,
        Weiwei Wan,~\IEEEmembership{Senior~Member,~IEEE},\\
        Hesheng Wang,~\IEEEmembership{Senior~Member,~IEEE},
        Kensuke Harada,~\IEEEmembership{Fellow,~IEEE}
\thanks{This work was conducted while Mengxin Xu was a visiting researcher at Osaka University, Japan. It was partially supported by the Natural Science Foundation of China under Grant 62225309, 62073222, U21A20480 and U1913204. (\emph{Corresponding Author: Weiwei Wan and Hesheng Wang}.)}
\thanks{Mengxin Xu is with the Department of Automation, Shanghai Jiao Tong University, Shanghai 200240, China (e-mail: mengxin\underline{~}xu@sjtu.edu.cn).}
\thanks{Weiwei Wan and Kensuke Harada are with the Department of System Innovation, Graduate School of Engineering Science, Osaka University, Toyonaka, Osaka 560-0043, Japan (e-mail: wan@sys.es.osaka-u.ac.jp, harada@sys.es.osaka-u.ac.jp). }
\thanks{Hesheng Wang is with the Department of Automation, the Key Laboratory of System Control and Information Processing of Ministry of Education and the Shanghai Engineering Research Center of Intelligent Control and Management, Shanghai Jiao Tong University, Shanghai 200240, China (e-mail: wanghesheng@sjtu.edu.cn). }%
}
\markboth{Under Review by an IEEE Journal, 2024.}
{Xu \MakeLowercase{\textit{et al.}}: Direction-Constrained Hierarchical Control Framework for Smooth pHRI}
\maketitle
\maketitle

\begin{abstract}
This paper proposes a control method to address the physical Human-Robot Interaction (pHRI) challenge in the context of hierarchical tasks. A common approach to managing hierarchical tasks is Hierarchical Quadratic Programming (HQP), which, however, cannot be directly applied to human interaction due to its allowance of arbitrary velocity direction adjustments. To resolve this limitation, we introduce the concept of directional constraints and develop a direction-constrained optimization algorithm to handle the nonlinearities induced by these constraints. The algorithm solves two sub-problems, minimizing the error and minimizing the deviation angle, in parallel, and combines the results of the two sub-problems to produce a final optimal outcome. The mutual influence between these two sub-problems is analyzed to determine the best parameter for combination. Additionally, the velocity objective in our control framework is computed using a variable admittance controller. Traditional admittance control does not account for constraints. To address this issue, we propose a variable admittance control method to adjust control objectives dynamically. The method helps reduce the deviation between robot velocity and human intention at the constraint boundaries, thereby enhancing interaction efficiency. We evaluate the proposed method in scenarios where a human operator physically interacts with a 7-degree-of-freedom robotic arm. The results highlight the importance of incorporating directional constraints in pHRI for hierarchical tasks. Compared to existing methods, our approach generates smoother robotic trajectories during interaction while avoiding interaction delays at the constraint boundaries.
\end{abstract}


\IEEEpeerreviewmaketitle

\section{Introduction}

Recent advancements in physical Human-Robot Interaction (pHRI) have significantly improved robots' abilities to support individuals \cite{survey1}\cite{survey2}. For example, pHRI has shown promising results in tasks such as load transportation \cite{Transportation}, collaborative drawing \cite{drawing}, surface polishing \cite{polish}, assembly \cite{assembly}, rehabilitation \cite{Rehabilitation}, etc. It provides humans with an intuitive method of engagement, allowing them to leverage their task-specific expertise without needing detailed knowledge of robot control. In pHRI, the robot can reduce both the physical and cognitive load on humans, while humans contribute valuable guidance based on their experience.

In many pHRI applications, the primary objective is to reduce human effort by requiring compliant behavior from the robot. However, in tasks that involve cognitive demands, such as assembly, incorporating task information is required to reduce the human's cognitive load and physical effort \cite{partial1}. The task information is interpreted as active constraints \cite{ac} during task execution. Robots shall not only respond compliantly to human forces but also satisfy these constraints to ease operational difficulty. Moreover, safety limits, such as the robot’s joint velocity, must also be considered during interaction. The various information and limits related to tasks are typically assigned to different task levels to form a hierarchical control problem. Conventionally, such a problem can be solved using the Hierarchical Quadratic Programming (HQP) \cite{HQP} method. However, a key drawback of existing HQP methods is that they allow arbitrarily adjusting robot motion directions. Directly applying HQP methods with insufficient motion direction adjustment to pHRI will lead to conflict with human operators and decrease user satisfaction. 

This article presents a new formulation and solution for pHRI in the context of hierarchical tasks. It focuses on scenarios where hierarchical constraints take precedence over interaction tasks. To ensure a reasonable robot motion direction, the direction deviations of the various tasks are modeled as a constraint and the problem is formulated as a direction-constrained optimization problem. Directly solving this problem is difficult as direction constraints are nonlinear and challenging to handle using existing HQP solvers. To address this, we propose a direction-constrained hierarchical optimization algorithm. The algorithm solves two sub-problems, minimizing the error and minimizing the deviation angle, in parallel, and combines the results of the two sub-problems under the direction constraints to produce a final optimal outcome. The proposed algorithm is an iterative process inspired by the active-set method for HQP \cite{hi4}.

Besides the optimization algorithm, the proposed control framework considers the influence of the constraints on the admittance control and leverages a variable admittance control method to reduce velocity deviation when the robot's motion is impeded by constraints. Admittance control \cite{ImpedanceControl} is the first step to convert human forces in pHRI. It helps determine the velocity objectives for complying with human motion. In admittance control, robot dynamics is typically modeled as a mass-spring-damper system, allowing it to adapt its behavior based on external forces from the human side. However, traditional admittance control does not account for constraints and may significantly reduce human-robot interaction efficiency at the constraint boundaries. To solve this problem, we propose a variable admittance control method to dynamically adjust control objectives at constraint boundaries. The method helps reduce the deviation between robot velocity and human intention and improves interaction efficiency. 

In summary, this work have the following contributions:
\begin{itemize}
\item[(1)] Proposal of hierarchical control framework that combines variable admittance control and optimization control with direction constraints.
\item[(2)] Proposal of an analytical algorithm to solve the direction-constrained hierarchical optimization problem.
\item[(3)] Design of a variable admittance controller to improve interaction performance when the robot's motion is constrained.
\end{itemize}

To evaluate the proposed framework, we conducted several experiments on pHRI tasks in the context of hierarchical tasks. The results showed that our method improves human comfort and interaction performance efficiency compared to HQP \cite{hi4} and task scaling methods \cite{SNS, eSNS}. We also examined the impact of variable admittance through ablation experiments. The results highlighted its advantages. Finally, we conducted a human-robot co-assembly task with three levels of priority and
demonstrated the effectiveness of our proposed method in the context of a practical application.

The rest of this article is organized as follows. Section \ref{Related Work} reviews related work. Section \ref{Preliminaries} presents preliminaries. Section \ref{Hieararchy} analyzes the hierarchical control problem in pHRI and formulates the direction-constrained problem. Section \ref{Algorithm} presents our propose direction-constrained optimization algorithm. Section \ref{Variable} shows the variable admittance design. Section \ref{Experiment} details the experiments and comparisons. Section \ref{Conclusion} concludes the paper.
\section{Related Work} \label{Related Work}
\subsection{pHRI under Hierarchical Tasks} 
Depending on the application requirements, a robot may need to perform a variety of tasks. When pHRI is the sole task, the control strategy focuses on closely following human intentions, as demonstrated in \cite{tii, Bayesian, Intent-Estimation}. In cases where additional tasks are required, the robot can be assigned multiple sub-goals alongside pHRI. The concept of assigning sub-goals was first introduced as ``virtual fixtures'' in telerobotic manipulation studies \cite{vf1}. Virtual fixtures, also known as active constraints, have since gained widespread use in fields such as robotic surgery and industrial automation. For instance, in \cite{vf2}, the movement of a robotic tool is constrained within a predefined range through the use of hard and soft virtual fixtures. Similarly, Papageorgiou et al. \cite{partial1} presented a passive pHRI control scheme with virtual constraints to assist in a folding assembly task. In this case, the sub-goal is defined within a subspace of $SE(3)$ and represented by parametric expressions. This work was further extended in \cite{partial2}, where artificial potentials were employed to implement the task geometry-based visual fixture. 

When multiple tasks need to be prioritized based on their importance, the virtual fixtures approach becomes less effective. For this reason,  many studies formulate pHRI with hierarchical tasks as a hierarchical control problem. Hierarchical control conventionally aims to manage multiple tasks simultaneously for redundant robots \cite{hi6}. In \cite{hi1} and \cite{hi2}, hierarchical impedance controllers were proposed, where subtasks were dynamically decoupled using dynamically consistent null space projectors. A passive task-prioritized shared-control method for teleoperation of redundant robots was presented in \cite{low-level}. However, these methods do not account for inequality constraints. To address this limitation, Hoffman et al. \cite{QP1} introduced a prioritized Cartesian impedance control framework with a cascade QP optimization to explicitly consider inequality constraints. For more intuitive reactions to external forces, Osorio et al. \cite{HQP} proposed a hierarchical framework combining null-space projection and QP-based redundancy solvers. In the framework, tasks are assigned different priority levels by incorporating a task-scaling factor in the QP problem. However, the QP solution does not guarantee smoothness in directional adjustments when constraints become active.

In this paper, we formulate the pHRI under hierarchical tasks problem as a hierarchical optimization problem with direction constraints. The tasks are modeled with both equality and inequality constraints, with pHRI set as the lowest-priority task. The directional constraints ensure that the robot moves within an acceptable range of directions. This formulation explicitly accounts for the deviation between the robot's velocity and human intention at the constraint boundaries. To the best of our knowledge, this work is the first to consider direction constraints in pHRI within the context of hierarchical tasks.

\subsection{Hierarchical Optimization with Directional Constraints}\label{2.2}
In early studies, tasks assigned to robots were defined by a set of equality constraints and solved by exploiting null space projection \cite{hi6}. Later research expanded to address both equality and inequality constraints. For example, Mansard et al. \cite{SoT} proposed a hierarchy-based control scheme for unilateral constraints, utilizing a potential field. More recently, Hierarchical Quadratic Programming (HQP) has been extensively investigated. Kanoun et al. \cite{hi3} presented a prioritized task regulation framework based on a sequence of QPs. Escande et al. \cite{hi4} presented a hierarchical solver based on complete orthogonal decomposition and active-set, formulating QP problems as sub-problems considering only equality constraints. Infeasible constraints were relaxed by using slack variables. However, the robot's behavior in the operational space can be adjusted arbitrarily since there are no directional constraints on the slack variables.

The task scaling technique \cite{scale} is commonly utilized to preserve task direction under constraints. When inequality constraints are violated, the speed of equality tasks is reduced to recover feasibility. The technique was combined with null space projections in \cite{SNS}, where the Saturation in the Null Space (SNS) algorithm was proposed. The SNS algorithm was further developed to handle constraints in both the joint and Cartesian spaces in \cite{SNS2}. Building upon \cite{SNS}, several enhanced versions were proposed in \cite{eSNS}, facilitating hierarchical redundancy resolution under arbitrary constraints for any low-level control type. Although these methods can maintain the task direction, they are not suitable for hierarchical optimization problems in the context of pHRI as they sacrifice flexibility in adjustable directions. A negative consequence of such sacrificing is that if the velocity of one direction in Cartesian space is constrained to zero, the scaling factor of the interaction task also becomes zero. In such cases, the robot will stop and be unable to move in any other direction.

In this paper, we propose a direction-constrained formulation for pHRI under hierarchical tasks. This formulation strikes a balance between the slack variable method and the scaling technique. By allowing a range of possible directions for robotic motion adjustments, it achieves a more human-friendly compromise. However, since the direction constraints are highly nonlinear and difficult for existing methods to handle, we propose a direction-constrained hierarchical optimization algorithm that minimizes control error and angle deviation in parallel and combines the results of the two minimizations to produce a final optimal outcome. 

\subsection{Variable Admittance Control}
In pHRI, compliant robot behavior is traditionally achieved by designing a desired relationship between force and position. This is typically realized by admittance control. The performance of an admittance controller is determined by its parameters like mass, damping, and stiffness coefficients.

Variable admittance control \cite{va-survey} allows adaptive changes to these parameters depending on different situations. For example, in \cite{var1}, the damping term was designed to vary with a task-related cost function. To improve system intuitiveness, the authors in \cite{Variable-admittance} proposed a variable admittance based on human intentions, including desired velocity and acceleration. In \cite{Variable-Impedance}, the damping varied according to the end-effector's velocity, recommending higher damping for smaller velocities. Similarly, to resist disturbances, Chen et al. \cite{sfc} designed a shear-thickening fluid controller that also features velocity-related damping. Additionally, some studies have adapted the parameters using learning-based methods. In \cite{rl}, reinforcement learning was used to learn appropriate damping to minimize trajectory jerk. In \cite{rl2}, integral reinforcement learning was combined with a linear quadratic regulator to optimize the impedance model.

The aforementioned literature primarily focuses on improving admittance performance based on human intention. In contrast, this paper presents a variable admittance that accounts for boundary conditions. The goal is to minimize the deviation between the velocity objective obtained through the admittance controller and the actual velocity of the robot. 
\section{Preliminaries}\label{Preliminaries}

\subsection{Robotic Tasks and Constraints}
Typical robotic tasks include moving along a straight line or rotating around a specific point. 
According to task requirements, certain equality and inequality constraints can be formulated. In this subsection, we present the preliminary formulations of these constraints. 

Let $\bm{x} \in \mathbb{R}^{m}$ denote a task vector in an $m$-dimensional task space and $\bm{q} \in \mathbb{R}^{n}$ denote the robot joint angles. Their relationship can be denoted by
\begin{equation}
    \dot{\bm{x}} = \boldsymbol{\mathcal{J} }\dot{\bm{q}},
\end{equation} 
where $\bmc{J}$ is a general task-related Jacobian matrix. 
To track a desired task trajectory $\bm{x}_d$, a reference velocity is typically provided by a first-order control law as
\begin{equation}
\label{eq-v}
    \bm{v} = \dot{{\bm{x}}}_d + \bmc{K} (\bm{x}_d - \bm{x}),
\end{equation}
where $\bmc{K}$ is a positive definite gain matrix. The control command of robot joint velocities $\bm{u}$ for tracking $\bm{y}_d$ shall meet the following equation
\begin{equation}
    \bmc{J}^{eq}\bm{u}=\bm{v},
    \label{eq_u_goal}
\end{equation}
where $\bmc{J}^{eq}$ is Jacobian matrix for equality task constraints.

On the other hand, if $\bm{y}$ is limited by a lower bound $\underline{\bm{y}}$ and an upper bound $\overline{\bm{y}}$, we can have the following velocity constraints using the shaping method described in \cite{eSNS}
\begin{equation}
\label{constraint}
\left\{
\begin{aligned}
       &\underline{\bm{v}}  = \dot{\underline{\bm{x}}} + \bmc{K} (\underline{\bm{x}} - \bm{x}) \\
   &\overline{\bm{v}}  = \dot{\overline{\bm{x}}} + \bmc{K} (\overline{\bm{x}} - \bm{x}).
\end{aligned}
\right.
\end{equation}
Here, $\underline{\bm{v}}$ and $\overline{\bm{v}}$ represent the lower and upper bounds for the task velocity $\bm{v}$, respectively. The control command for the robot joint velocities must satisfy the following boundary conditions
\begin{equation}
    \underline{\bm{v}} \leq \bmc{J}^{iq}\bm{u} \leq \overline{\bm{v}}.
    \label{eq_u_bounds}
\end{equation}
Here, $\bmc{J}^{iq}$ is Jacobian matrix for inequality task constraints.

Equations \eqref{eq_u_goal} and \eqref{eq_u_bounds} are usually written in a general form as
\begin{equation}
    \bm{A} \bm{u} = \bm{b}, ~~\bm{C} \bm{u} \leq \bm{d},
    \label{eq_con}
\end{equation}
where $\bm{A}$ replaces $\bmc{J}^{eq}$ to denote the task Jacobian matrix of the equality constraint. $\bm{b}$ replaces $\bm{v}$ to denote the velocity reference. $\bm{u}$ remains to be the control command of the joint velocities. $\bm{C}$ and $\bm{d}$ are new variables introduced for unifying the bounds, where
\begin{equation}
\left\{
\begin{aligned}
& \bm{C}= \begin{bmatrix}
    -\bmc{J}^{iq} & \bmc{J}^{iq}
\end{bmatrix}^\top\\
& \bm{d} = \begin{bmatrix}
    -\underline{\bm{v}} & \overline{\bm{v}}
\end{bmatrix}^\top.
\end{aligned}\right.
\end{equation}

In case of multiple tasks, each variable mentioned above will be indexed to indicate the task level as follows except the variable $\bm{u}$ since it is responsible for all levels.
\begin{equation}
\label{con}
    \bm{A}_k \bm{u} = \bm{b}_k, ~~\bm{C}_{k} \bm{u} \leq \bm{d}_k.
\end{equation}

\subsection{Interaction Task considering Admittance Control}
For the human-robot interaction tasks, we use admittance control to obtain the velocity objectives for complying with human motion. Admittance control is widely employed to control a robot manipulator compliantly in the presence of human force guidance. The conventional admittance model is expressed as
\begin{equation}
\label{ad}
    \bm{M}(\Ddot{\bm x}_a^\mathsf{E}-\Ddot{\bm x}_d^\mathsf{E}) + \bm{D}(\Dot{\bm x}_a^\mathsf{E}-\Dot{\bm x}_d^\mathsf{E}) + \bm{K} (\bm{x}_a^\mathsf{E} - \bm{x}_d^\mathsf{E}) = \bm{f}_{ext},
\end{equation}
where $\bm{M}$, $\bm{D}$, and $\bm{K}$ are matrices representing the desired mass, damping, and stiffness coefficients, respectively. These matrices are typically chosen to be diagonal and positive. $\bm{x}^\mathsf{E}$ is specifically the states of robot's end-effector containing positions and orientations, and $\bm{x}_d^\mathsf{E}$ is the trajectory desired by a subtask. $\bm{x}_a^\mathsf{E}$ denotes the position objective computed using the admittance controller. $\bm{f}_{ext}$ denotes the external force acted on the robot by human operators.

In cases where the robot is entirely compliant and follows the human without a predefined trajectory, the stiffness term will be removed, and equation \eqref{ad} is simplified to
\begin{equation}
    \label{ad2}
    \bm{M} \dot{\bm v}_a+ \bm{D}  \bm{v}_a = \bm{f}_{ext},
\end{equation}
where $\bm{v}_a = \dot{\bm{x}}_a^\mathsf{E}$ denotes the velocity objective of the end-effector computed from the admittance controller. 

Once we obtain $\bm{v}_a$, the joint velocity command $\bm{u}$ can be formulated accordingly as
\begin{equation}
\label{eq-u-va}
    \bm{J}^\mathsf{E} \bm{u} = \bm{v}_a,
\end{equation}
where $\bm{J}^\mathsf{E}$ represents the Jacobian matrix of the end-effector. We can also formulate the velocity constraints for the interaction task as 
\begin{equation}
\label{constraint2}
\left\{
\begin{aligned}
       &\underline{\bm{v}}_a  = \dot{\underline{\bm{x}}}^\mathsf{E} + \bmc{K} (\underline{\bm{x}}^\mathsf{E} - \bm{x}^\mathsf{E}) \\
   &\overline{\bm{v}}_a  = \dot{\overline{\bm{x}}}^\mathsf{E} + \bmc{K} (\overline{\bm{x}}^\mathsf{E} - \bm{x}^\mathsf{E}).
\end{aligned}
\right.
\end{equation}

Note that in the above formulation, we use \eqref{ad2} instead of \eqref{eq-v} to compute reference velocity and achieve compliant robot behavior in the interaction task. However, \eqref{constraint2} has the same form as \eqref{constraint} since the constraints are independent of the interaction. 

Using \eqref{eq-u-va} and \eqref{constraint2}, we can formulate the control constraints of an interaction task in the same form as \eqref{eq_con}. The constraints can be represented integrally with the robotic tasks and constraints using $\bm{A}_k \bm{u} = \bm{b}_k, \bm{C}_{k} \bm{u} \leq \bm{d}_k$.

% ...


\subsection{Multi-Task pHRI Problem}

The multi-task pHRI problem, considering the multiple equality and inequality constraints at all task levels, can be formulated as
\begin{equation}
\setlength{\arraycolsep}{0.3pt}
\label{op1}
\begin{aligned}
    &\quad \quad \quad  \mathop{\mathrm{lex} \min}_{\bm{u},\bm{w}_1, \cdots \bm{w}_N}  \{||\bm{w}_1||^2 \cdots ||\bm{w}_N||^2\}\\
    &\mathrm{s.t.} \\
    &\begin{array}{llll}
        ~ & \mathrm{level~1} 
        \mathrm{(high~priorty)}: & ~~\bm{A}_1 \bm{u} = \bm{b}_1 + \bm{w}_1, & ~~\bm{C}_{1} \bm{u} \leq \bm{d}_1,\\
        ~ & \quad\vdots  \\
        ~ & \mathrm{level~N} 
        \mathrm{(low~priorty)}: & ~~\bm{A}_N \bm{u} = \bm{b}_N + \bm{w}_N, & ~~\bm{C}_{N} \bm{u} \leq \bm{d}_N.\\
    \end{array}\\
\end{aligned}
\end{equation}
Here, $\mathrm{lex}$ indicates the optimization problem is a lexicographic one that prioritizes constraints according to their task levels. The solution process first involves satisfying the constraints at task levels with high priority and, once these are met, seeks a multi-objective optimal solution that complies with the low-priority constraints. $\bm{w}_k$ is a slack variable \cite{slack} for the equality constraints. It is used to relax the task reference trajectory when there are conflicts with inequality constraints. Typically, the primary tasks that a robot is expected to accomplish are assigned with high priority, while tasks related to accommodating human interaction forces are assigned to task levels with low priority \cite{low-level}. This hierarchical structure allows the optimization solver to prioritize the resolution of the primary tasks.

When only considering the equality constraints in (\ref{op1}), the least-norm optimal solution can be computed recursively for $N$ levels following \cite{Siciliano}
\begin{equation}
\label{eq}
\begin{aligned}
    &\bm{u}_k = \bm{u}_{k-1} + (\bm{A}_k \bm{P}_{k-1})^\dagger (\bm{b}_k - \bm{A}_k \bm{u}_{k-1})\\
    &\bm{u} = \bm{u}_N,
\end{aligned}
\end{equation}
where $(\bm{\cdot})^\dagger$ represents the Moore–Penrose pseudo-inverse of a matrix, $\bm{u}_0 = 0$, $\bm{P}_0 = \bm{I}_{n \times n}$, and $\bm{P}_k \in \mathbb{R}^{n\times n}$ is the augmented projector into the null space of $[\bm{A}_1^T, \cdots, \bm{A}_k^T]^T$. $\bm{P}_k$ is computed recursively as \cite{SNS}:
\begin{equation}
    \bm{P}_k = \bm{P}_{k-1} - (\bm{A}_k \bm{P}_{k-1})^\dagger (\bm{A}_k \bm{P}_{k-1}).
\end{equation}
When considering both equality and inequality constraints, the optimal joint velocities can be solved using the HQP method, where each hierarchical level is defined as the following single optimization problem:
\begin{equation}
\begin{array}{lll}
    &\quad \quad \quad & \mathop{\min}_{\bm{u}_k, \bm{w}_k} ||\bm{w}_k|| \\
    &\mathrm{s.t.} &\bm{A}_{1 \rightarrow k-1} \bm{u}_k = \bm{A}_{1 \rightarrow k-1} \bm{u}_{k-1}\\
    && \bm{A}_k \bm{u}_{k}  = \bm{b}_k + \bm{w}_k \\
    && \bm{C}_{1 \rightarrow k} \bm{u}_k\leq \bm{d}_{1 \rightarrow k},
\end{array}
\end{equation}
where $\bm{A}_{1 \rightarrow k} = [\bm{A}_1^T, \cdots, \bm{A}_k^T]^T$, $\bm{C}_{1 \rightarrow k} = [\bm{C}_1^T, \cdots, \bm{C}_k^T]^T$ and $\bm{d}_{1 \rightarrow k} = [\bm{d}_1^T, \cdots, \bm{d}_k^T]^T$. Each single optimization can be solved using an existing method such as the active-set algorithm.

\subsection{Task Scaling Method for Hierarchical Optimization}

While solving \eqref{op1} using HQP can meet the required constraints, the resulting robot movements often lack smoothness. This is because the direction of $\bm{w}_k$ is unrestricted, allowing it to arbitrarily affect the behavior of the robot. To overcome this issue, researchers have proposed the task scaling method. In this approach, the term $\bm{b}_k+\bm{w}_k$ is transformed into the form $s\bm{b}_k$, where $s$ is a scaling factor. Essentially, $\bm{w}_k$ is redefined as $(s-1)\bm{b}_k$, thereby constraining its direction to be aligned with $\bm{b}_k$. The alignment ensures that the optimization preserves the task directions, leading to smoother and more coordinated robot behavior. Using the task scaling method, the optimization problem \eqref{op1} can be reformulated as follows:

\begin{equation}
\setlength{\arraycolsep}{0.3pt}
\label{op-scale}
\begin{aligned}
    &\quad  \mathop{\mathrm{lex} \min}_{\bm{u},s_1, \cdots s_N}  \{(1-s_1)^2 \cdots (1-s_N)^2\}\\
    &\mathrm{s.t.} \\
    &\begin{array}{llll}
        ~ & \mathrm{level} ~ 1\mathrm{(high~priority)}: & ~~\bm{A}_1 \bm{u} = s_1\bm{b}_1, & ~~\bm{C}_{1} \bm{u} \leq \bm{d}_1,\\
        ~ & \quad\vdots  \\
        ~ & \mathrm{level} ~ N\mathrm{(low~priority)}: & ~~\bm{A}_N \bm{u} = s_N\bm{b}_N, & ~~\bm{C}_{N} \bm{u} \leq \bm{d}_N.\\
    \end{array}\\
\end{aligned}
\end{equation}
Practical methods for solving \eqref{op-scale} are based on null space projections, as described in detail in \cite{SNS} and \cite{eSNS}.

The essence of task scaling is limiting the direction of $\bm{A}_k \bm{u}$ to be the same as the direction of $\bm{b}_k$. Therefore, the constraint at the $k$th level can also be written as follows:
\begin{equation}
\label{dc}
    \bm{A}_k \bm{u} = \bm{b}_k + \bm{w}_k, ~~\angle (\bm{A}_k \bm{u}, \bm{b}_k) = 0, ~~\bm{C}_{k} \bm{u} \leq \bm{d}_k,
\end{equation}
where $\angle (\bm{\cdot}, \bm{\cdot})$ represents the angle between two vectors.

\section{Direction-Constrained Formulation and Overview of the Proposed Method} \label{Hieararchy}

As discussed in the previous section, the pHRI problem involving multiple tasks can be formulated as either equations \eqref{op1} or \eqref{op-scale}. Solving the optimization problem in the form of equation \eqref{op1} allows the robot’s velocity to approximate $\bm{v}_a$ under hierarchical constraints. However, this formulation has a notable drawback in that the direction of the robot’s motion can be adjusted arbitrarily, leading to disjointed behavior and potentially reducing human comfort during interaction. In contrast, equation \eqref{op-scale} provides a more appropriate solution, as it imposes a well-defined robot behavior. Nevertheless, in human-robot cooperation scenarios, it is challenging for the human operator to apply force precisely in their intended direction. The uncertainty in force direction may cause $\bm{v}_a$ to violate the inequality constraints and lead to waiting for human adjustments at constraint boundaries. Moreover, since all components of the velocity vector are scaled uniformly, the components that may potentially satisfy constraints are also unnecessarily reduced.

In this work, we formulate a new problem by exploring an intermediate model between formulations (12) and (16). The core of this formulation is to relax the direction constraint $\angle (\bm{A}_k \bm{u}, \bm{b}_k) = 0$ in \eqref{dc} to allow a range of possible directions, thus balancing interaction comfort and cooperation efficiency. Equation \eqref{op-scale} is the mathematical representation of the new problem:
\begin{equation}
\setlength{\arraycolsep}{0.4pt}
\label{op2}
\begin{aligned}
    &\quad \quad \quad  \mathop{\mathrm{lex} \min}_{\bm{u},\bm{w}_1, \cdots \bm{w}_N}  \{||\bm{w}_1||^2 \cdots ||\bm{w}_N||^2\}\\
    &\mathrm{s.t.} \\
 &\begin{array}{llll}
         ~ &\bm{A}_1 \bm{u} = \bm{b}_1 + \bm{w}_1, &~~\angle (\bm{A}_1 \bm{u}, \bm{b}_1) \leq \theta_1, &~~\bm{C}_{1} \bm{u} \leq \bm{d}_1\\
        ~ & \quad \quad \quad \quad \vdots  &\quad \quad \quad \quad \vdots &\quad \quad \quad  \vdots\\
         ~ &\bm{A}_k \bm{u} = \bm{b}_k + \bm{w}_k, &~~\angle (\bm{A}_k \bm{u}, \bm{b}_k) \leq \theta_k, &~~\bm{C}_{k} \bm{u} \leq \bm{d}_k\\
         ~ &\quad \quad \quad \quad \vdots  &\quad \quad \quad \quad \vdots &\quad \quad \quad \vdots\\
         ~ &\underbrace{\bm{A}_N \bm{u} = \bm{b}_N + \bm{w}_N}, &~~\underbrace{\angle (\bm{A}_N \bm{u}, \bm{b}_N) \leq \theta_N}, &~~ \underbrace{\bm{C}_{N} \bm{u} \leq \bm{d}_N}.\\
          ~ & \text{\small{Task equality}} &~~~~\text{\small{Direction constraints}} &\text{\small{Task inequality} }\\
          ~ &\text{\small (t-eq) constraints } &~ &\text{\small (t-iq) constraints}
    \end{array}
\end{aligned}
\end{equation}
\begin{figure}[!htbp]
    \centering
    \includegraphics[width=0.42\textwidth]{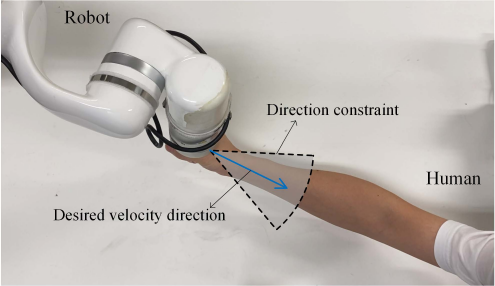}
    \caption{Direction constraints in pHRI.}
    \label{problem}
\end{figure}

Here, $\angle (\bm{A}_k \bm{u}, \bm{b}_k) = 0$ is relaxed to $\angle (\bm{A}_k \bm{u}, \bm{b}_k) \leq \theta_k(t)$ to allow more flexibility. At each level, there are three constraints: task equality (t-eq) constraints, direction constraints, and task inequality (t-iq) constraints. The angle $\theta_k$ ranges from $0$ to $\pi$, with $\theta_k=0$ meaning the task direction should be the same as the original one, and $\theta_k=\pi$ meaning the task direction can be arbitrary. The equation \eqref{op2} is a more general form compared to \eqref{op-scale} where the explicit physical meaning of the $\theta_k$ constraints should be guaranteed in practice. When we let $\theta_k$ of all levels equal zero, \eqref{op2} becomes \eqref{op-scale}. 
The basic concept of the direction constraints is illustrated in Fig. \ref{problem}. Specifically, the relaxed constraints ensure that the robot’s velocity does not significantly deviate from its original direction. The choice of constraint bounds depends on the conditions of the task at hand. For instance, in tasks requiring delicate operations, the velocity direction should closely align with the original, thus a narrower range is assigned. In contrast, for tasks where rougher movements are acceptable, a wider range of directions can be permitted.

\begin{figure}[!htbp]
    \centering
    \includegraphics[width=\linewidth]{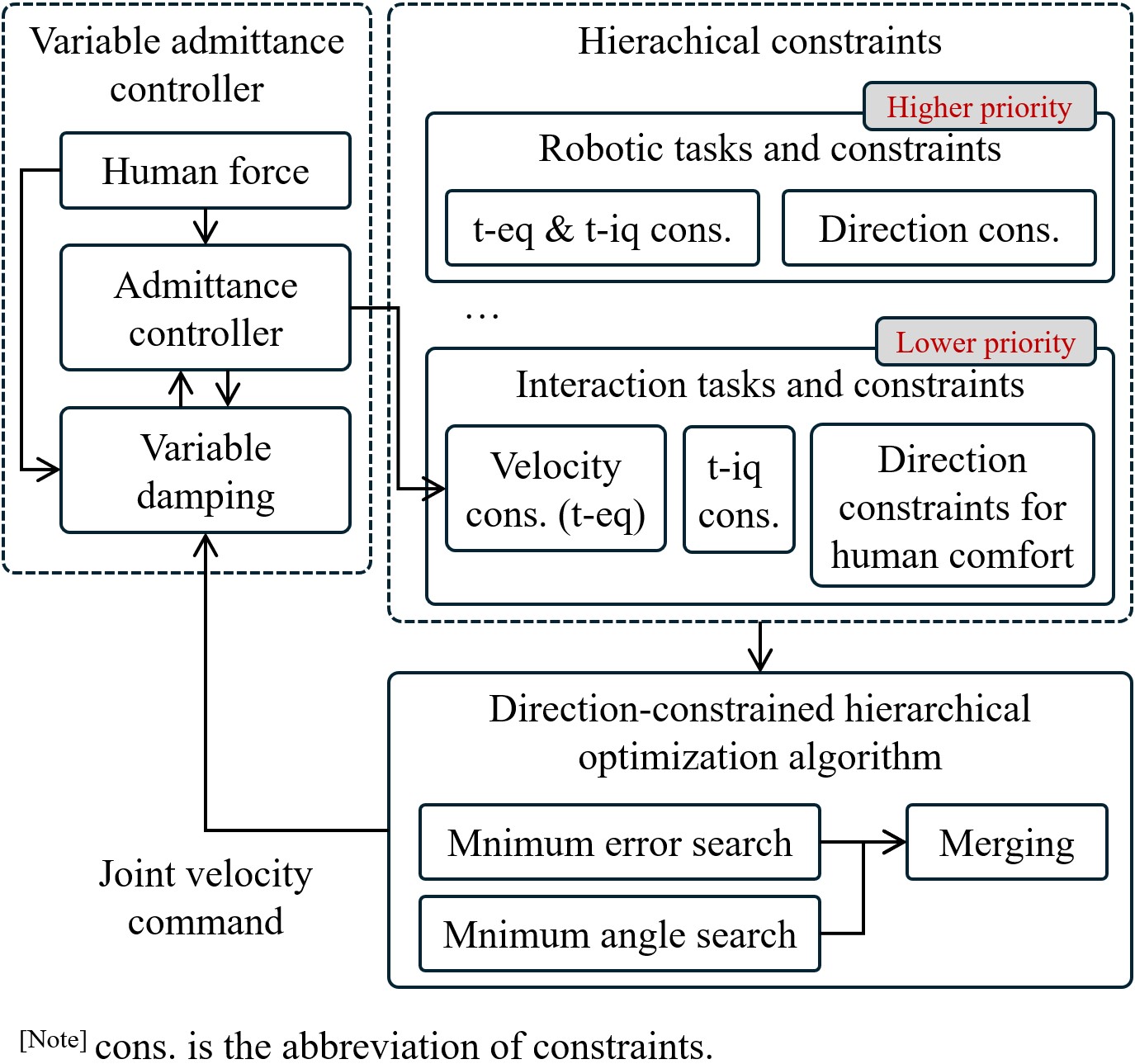}
    \caption{Overall control framework.}
    \label{framework}
\end{figure}

Note that the direction constraint can be expressed using the vector cosine formula as follows:
\begin{equation}
    \frac{(\bm{A}_k \bm{u})^T \bm{b}_k}{||\bm{A}_k \bm{u}|| \cdot ||\bm{b}_k||} \geq \cos \theta_k.
\end{equation}
This constraint is highly nonlinear because of the vector magnitude term $||\bm{A}_k \bm{u}||$, making it difficult for existing methods to solve. In this paper, we develop an algorithm to efficiently find a solution that meets these constraints. The overall control framework of our method is shown in Fig. \ref{framework}. At the beginning of each control loop, the admittance controller computes a desired velocity for the end-effector based on the human force. Variable damping is accounted in the process for cases where the robot motion is limited by constraints (see Section \ref{Variable}). The computed desired velocity is set as a subtask with the lowest priority, while prior task knowledge forms the subtasks as t-eq and t-iq constraints at higher priority levels. The t-eq and t-iq constraints are then combined with direction constraints to form the problem (\ref{op2}). This problem is solved by the developed direction-constrained optimization algorithm, which provides the joint velocity command. The deviation between the velocity objective obtained through the admittance controller and the velocity of the robot is used to update the damping term. Finally, the admittance controller recomputes the desired velocity for the next control loop. The details of this framework, along with the specific optimization algorithm, will be presented in the following sections.

\section{Direction-Constrained Hierarchical Optimization Algorithm} \label{Algorithm}




\begin{figure*}[h]
    \centering
    \includegraphics[width=\textwidth]{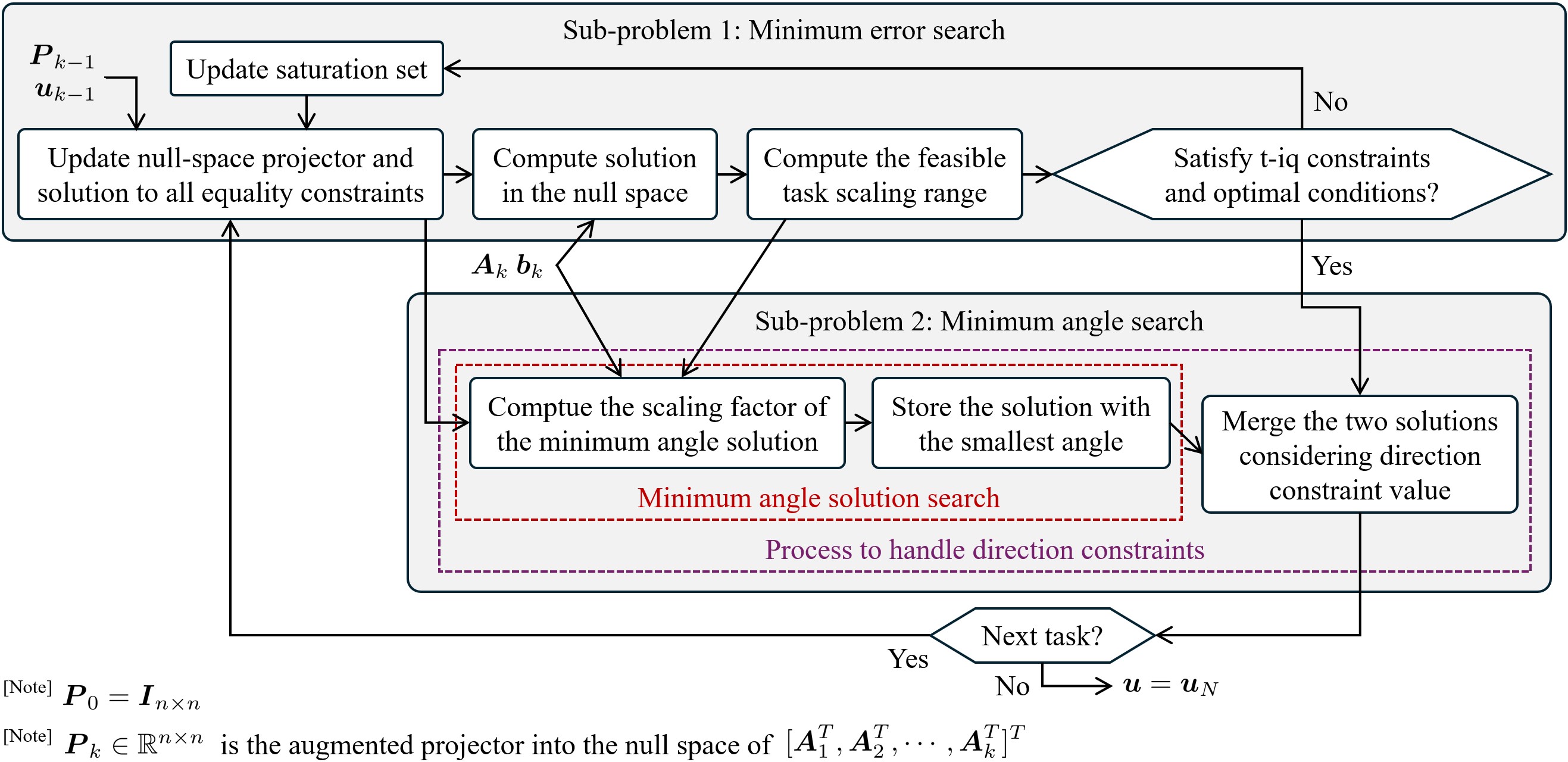}
    \caption{Workflow of the proposed direction-constrained hierarchical optimization algorithm.}
    \label{workflow}
\end{figure*}

The proposed algorithm recursively calculates the control variables at each level from high priority to low priority. Due to the nonlinear characteristics of direction constraints, considering all kinds of constraints of \eqref{op2} simultaneously is difficult. Therefore, we split \eqref{op2} into two sub-problems to solve. The first sub-problem is similar to \eqref{op1}, which minimizes the task error without considering task directions. The optimization goal of the first sub-problem is expressed as
\begin{equation}
    \mathop{\mathrm{lex} \min}_{\bm{u}}  \{||\bm{A}_1 \bm{u}-\bm{b}_1|| \cdots ||\bm{A}_N \bm{u}-\bm{b}_N||\}.
\end{equation}
The second sub-problem focuses on minimizing the task angle expressed as
\begin{equation}
\label{sub-op2}
    \mathop{\mathrm{lex} \min}_{\bm{u}}  \{\angle (\bm{A}_1 \bm{u}, \bm{b}_1) \cdots \angle (\bm{A}_N \bm{u}, \bm{b}_N)\}.
\end{equation}
The t-iq constraints should be satisfied in both sub-problems. However, solving \eqref{sub-op2} under t-iq constraints separately is challenging. Therefore, we solve the second sub-problem in parallel with the first one while using the intermediate results at each iteration. 

We refer to the processes of solving these two sub-problems as the minimum error search and the minimum angle search, respectively. The solutions to these sub-problems are denoted by $\bm{\mu}_1$ and $\bm{\mu}_2$. The core strategy for obtaining a feasible solution to the problem (\ref{op2}) is to identify the optimal combination of these two solutions. Here, we choose to merge the solutions of these sub-problems through a linear combination considering that they must both meet the i-eq constraints. We prove that the result of this linear combination exhibits a monotonic relationship with respect to the parameter settings, and we leverage this relationship to obtain the optimal merging outcome. The overall workflow of the proposed algorithm is presented in Fig. \ref{workflow}, and the details are provided in pseudocode form as Algorithm \ref{al1}.

\begin{algorithm}[h]
    \caption{Direction-constrained hierarchical optimization}
    \label{al1}
    \begin{algorithmic}[1]
    \State $\bm{P}_0 = \bm{I}, \bm{u}_0 = 0$
    \For{$k=1$ to N}
        \State $s^* = 0, \alpha^*=\pi$
        \State $\mathcal{S} =  \mathrm{Null}$
        \State $\widetilde{\bm{P}}_k = \bm{P}_{k-1}$
        \State $\widetilde{\bm{u}}_{k} = \bm{u}_{k-1}$
        \Repeat
            \State optimal = True
            \State $\widetilde{\bm{A}}_k = \bm{A}_k \widetilde{\bm{P}}_k$
            \State $\bm{\mu}_1 = \widetilde{\bm{u}}_{k} + \widetilde{\bm{A}}_k^\dagger (\bm{b}_k - \bm{A}_k \widetilde{\bm{u}}_{k} )$
            \State $\bm{z} = \bm{C}_{1 \rightarrow k} \widetilde{\bm{A}}_k^\dagger \bm{b}_k$
            \State $\bm{r} = \bm{C}_{1 \rightarrow k} \bm{\mu}_1 - \bm{z}$
            \State $s_{min},s_{max},j=$ getRange($\bm{z},\bm{r}$) \Comment{call Alg. 2}

            \State $\bm{\gamma} = \bm{b_k} - \bm{A}_k \widetilde{\bm{A}}_k^{\dagger}\bm{b_k}$
            \State $\bm{\epsilon} = \bm{A}_k \widetilde{\bm{u}}_{k} - \bm{A}_k \widetilde{\bm{A}}_k^{\dagger} \bm{A}_k \widetilde{\bm{u}}_{k}$

            \If{$\bm{\gamma}=\bm{0}$ or $\bm{\epsilon}=\bm{0}$}
                \State $s = 1$
            \Else
                \State $s = $ max($(\bm{\epsilon}^T \bm{\epsilon})/(\bm{\epsilon}^T \bm{\gamma})$, ~0)
            \EndIf
            
            \State $s = $ max(min($s$, $s_{max}$),$s_{min}$)
            
            \State $\bm{\mu}_2^{\prime} = \widetilde{\bm{u}}_{k} + \widetilde{\bm{A}}_k^{\dagger}(s\bm{b}_k - \bm{A} \widetilde{\bm{u}}_{k} )$

            \State $\alpha = \angle(\bm{A} \bm{\mu}_2^{\prime}, \bm{b}_k)$

            \If{$\alpha < \alpha^*$\textbf{or} ($\alpha = \alpha^*$ \textbf{and} $||s-1|| < ||s^*-1||$)}

                \State $\bm{\mu}_2 = \bm{\mu}_2^\prime, \alpha^*=\alpha, s^*=s$
            \EndIf

            \If{$s_{max}<1$ \textbf{or} $s_{min} > 1$}

                \State $\mathcal{S} \gets \mathcal{S} \cup \{j\} $
                \State optimal = False
    
            \Else
                \State $\bm{\lambda} = -(\bm{A}_k \bm{P}_{k-1}\widetilde{\bm{C}}_{sat}^\dagger)^T(\bm{A}_k\bm{\mu}_1-\bm{b}_k)$ 

                \State $\lambda_{min}, index =$ min\{$\bm{\lambda}$\} 

                \If{$\lambda_{min} < 0$}
                    \State $\mathcal{S} \gets \mathcal{S} \backslash \{\mathcal{S}_{index}\} $
                    \State optimal = False
                \EndIf          
            \EndIf

            \If{optimal = False}
                \State update $\bm{C}_{sat}$ and $\bm{d}_{sat}$

                \State $\widetilde{\bm{C}}_{sat} = {\bm{C}}_{sat} \bm{P}_{k-1}$

                \State $\widetilde{\bm{P}}_k = (\bm{I} - \widetilde{\bm{C}}_{sat}^\dagger {\bm{C}}_{sat}) \bm{P}_{k-1}$

                \State $\widetilde{\bm{u}}_k = \bm{u}_{k-1} + \widetilde{\bm{C}}_{sat}^\dagger (\bm{d}_{sat} - \bm{C}_{sat} \bm{u}_{k-1})$
            \EndIf
        \Until{optimal = True}
        \State $\bm{u}_k = $ merge($\bm{\mu}_1,\bm{\mu}_2, \theta_k$) \Comment{call Alg. 3}
        \State $\bm{P}_k = \bm{P}_{k-1} - (\bm{A}_k \bm{P}_{k-1})^\dagger (\bm{A}_k \bm{P}_{k-1})$
    \EndFor
    \State $\bm{u} = \bm{u}_N$
    \end{algorithmic}
\end{algorithm}

\begin{algorithm}[h]
    \caption{Scaling range computation}
    \label{al2}
    \begin{algorithmic}[1]
    \Function{getRange}{$\bm{z},\bm{r}$}
    \State $\underline{\mathcal{S}} = \overline{\mathcal{S}}= $ Null
    \For{$i=1$ to $l_k$}
       \State \quad $s_i = (d_i - r_i)/z_i$
        \If{$z_i< 0$}
            \State $\underline{\mathcal{S}} \gets \underline{\mathcal{S}} \cup \{(s_i, i)\}$
        \Else
            \State $\overline{\mathcal{S}} \gets \overline{\mathcal{S}} \cup \{(s_i, i)\}$
        \EndIf
    \EndFor
    \State $s_{max}, j_1 = $ min\{$\overline{\mathcal{S}}$\}
    \State $s_{min}, j_2 = $ max\{$\underline{\mathcal{S}}$\}
    \State most critical constraint $= j_1$
    \If{$s_{min}>s_{max}$ \textbf{or} $s_{max} < 0$}
        \State $s_{min}=s_{max}=0$
    \EndIf
    \Return $s_{min}, s_{max}, j_1$
    \EndFunction
    \end{algorithmic}
\end{algorithm}

\subsection{Minimum Error Search}
The hierarchical active-set method is adopted to obtain the solution $\bm{\mu}_1$ for the first sub-problem. This method is an iterative search process. During the process, $\bm{\mu}_1$ is computed in the null space of all the equality constraints, similar to (\ref{eq})
\begin{equation}
\label{mu1}
    \bm{\mu}_1 = \widetilde{\bm{u}}_{k} + (\bm{A}_k \widetilde{\bm{P}}_k)^{\dagger}(\bm{b}_k - \bm{A} \widetilde{\bm{u}}_{k}),
\end{equation}
with the following iterative assignment
\begin{equation*}
    \widetilde{\bm{u}}_{k} = \bm{u}_{k-1}, \widetilde{\bm{P}}_k = \bm{P}_{k-1}.
\end{equation*}
Here, $\widetilde{\bm{u}}_{k}$ represents the control input that fulfills the tasks from level 1 to k-1 and the equality constraints converted from inequality constraints. $\widetilde{\bm{P}}_k$ denotes their null space projector. The update of these two variables will be detailed later.

Following the update of $\bm{\mu}_1$, the process checks whether $\bm{\mu}_1$ satisfies all t-eq constraints up to the current level. The checking is accomplished by computing a scaling range $[s_{min}, s_{max}]$ of a factor $s$ such that $s\bm{b}_k$ can be fulfilled under all the available t-iq constraints. The process to obtain $s_{min}$ and $s_{max}$ is outlined in Algorithm \ref{al2}. The input arguments for Algorithm \ref{al2} are given as
\begin{equation}
    \begin{aligned}
        &\bm{z} = \bm{C}_{1 \rightarrow k} (\bm{A}_k \widetilde{\bm{P}}_k)^\dagger \bm{b}_k\\
        &\bm{r} = \bm{C}_{1 \rightarrow k} \bm{\mu}_1 - \bm{z}.
    \end{aligned}
\end{equation}
If the obtained $[s_{min}, s_{max}]$ covers $1$, $\bm{\mu}_1$ is an admissible solution. Otherwise, the algorithm identifies that certain t-iq constraints are violated, and convert them into equality constraints. Specifically, the algorithm finds the most critical t-iq constraint up to the current level, which corresponds to the smallest scaling factor, converts it to equality constraint, and saves the converted equation into a saturation set $\mathcal{S}$. 

On the other hand, if \(\bm{\mu}_1\) satisfies all t-iq constraints up to the current level, its optimality is checked using the Karush-Kuhn-Tucker (KKT) criteria. This is implemented by determining whether any constraints in the current saturation set \(\mathcal{S}\) can be removed to find a better set. Here we write the control solution at the $k$th level in the following form to facilitate analysis:
\begin{equation}
    \bm{u}_k = \bm{u}_{k-1} + \bm{P}_{k-1} \bar{\bm{u}}_k.
\end{equation}
$\bm{P}_{k-1} \bar{\bm{u}}_k$ in the equation represents the null space of the previous tasks. The optimization problem at the \(k\)th level is thus equivalent to:
\begin{equation}
\begin{aligned}
    &\mathop{\min}_{\bar{\bm{u}}_k, \bm{w}_k} ||\bm{w}_k|| \\
    &\mathrm{s.t.} \quad \bm{A}_k(\bm{u}_{k-1} + \bm{P}_{k-1} \bar{\bm{u}}_k) = \bm{b}_k + \bm{w}_k \\
    &\quad \quad \quad \bm{C}_{\text{sat}} (\bm{u}_{k-1} + \bm{P}_{k-1} \bar{\bm{u}}_k) = \bm{d}_{\text{sat}}.
\end{aligned}
\end{equation}
The notation $\bm{C}_{sat}$ in the equation represents the rows of $\bm{C}_{1 \rightarrow k}$ whose indices are in $\mathcal{S}$, and the notation $\bm{d}_{sat}$ represents the corresponding rows of $\bm{d}_{1 \rightarrow k}$. The optimality conditions for this problem are:
\begin{subequations}
    \begin{equation}
        \bm{w}_k = \bm{A}_k \bm{P}_{k-1} \bar{\bm{u}}_k - (\bm{b}_k - \bm{A}_k \bm{u}_{k-1})
    \end{equation}
    \begin{equation}
        \bm{C}_{\text{sat}} \bm{P}_{k-1} \bar{\bm{u}}_k = \bm{d}_{\text{sat}} - \bm{C}_{\text{sat}} \bm{u}_{k-1}
    \end{equation}
    \begin{equation}
        (\bm{C}_{\text{sat}} \bm{P}_{k-1})^T \bm{\lambda} = - (\bm{A}_k \bm{P}_{k-1})^T \bm{w}_k,
    \end{equation}
\end{subequations}
where \(\bm{\lambda}\) is the Lagrange multiplier. Using the solution \(\bm{\mu}_1\), \(\bm{\lambda}\) can be computed as:
\begin{equation}
    \bm{\lambda} = -(\bm{A}_k \bm{P}_{k-1} (\bm{C}_{\text{sat}} \bm{P}_{k-1})^\dagger)^T (\bm{A}_k \bm{\mu}_1 - \bm{b}_k).
\end{equation}
If the current saturation set \(\mathcal{S}\) is optimal, the components of \(\bm{\lambda}\) should satisfy \(\lambda_i \geq 0\). Otherwise, we find the index of the minimum component of \(\bm{\lambda}\) and remove the component with the same index in \(\mathcal{S}\).

Through the above updating and checking process for $\bm{\mu}_1$, the saturation set \(\mathcal{S}\) may change by adding or removing a constraint. Once \(\mathcal{S}\) is changed, \(\bm{C}_{\text{sat}}\) and \(\bm{d}_{\text{sat}}\) will change accordingly. Then, \(\widetilde{\bm{P}}_{k}\) and  \(\widetilde{\bm{u}}_{k}\) will be recomputed as follows:
\begin{equation}
    \begin{aligned}
        &\widetilde{\bm{P}}_k = (\bm{I} - \widetilde{\bm{C}}_{\text{sat}}^\dagger {\bm{C}}_{\text{sat}}) \bm{P}_{k-1}, \\
        &\widetilde{\bm{u}}_k = \bm{u}_{k-1} + \widetilde{\bm{C}}_{\text{sat}}^\dagger (\bm{d}_{\text{sat}} - \bm{C}_{\text{sat}} \bm{u}_{k-1}),
    \end{aligned}
\end{equation}
with \(\widetilde{\bm{C}}_{\text{sat}} = {\bm{C}}_{\text{sat}} \bm{P}_{k-1}\). After that, the updating and checking process for $\bm{\mu}_1$ process will be performed again starting from equation \eqref{mu1} using the recomputed values. Conversely, if \(\mathcal{S}\) remains unchanged after the process, the iterations for the current level are finished and the algorithm moves forward to the merging stage.

\subsection{Minimum Angle Search}
Since the direction constraints are highly nonlinear, finding the global optimal solution that minimizes the task angle error is challenging. Instead of the global optimal solution, we find a locally optimal solution under the equality constraint set determined by the process of the previous subsection (t-eq and converted equality constraints). In other words, the process of finding the solution \(\bm{\mu}_2\) with the minimum angle is carried out in parallel with the updating and checking iterations of \(\bm{\mu}_1\). The computation of \(\bm{\mu}_2\) is based on the updated values of $\widetilde{\bm{P}}_k$ and $\widetilde{\bm{u}}_{k}$ obtained during solving \(\bm{\mu}_1\). 

Details of the minimum angle search are as follows. According to the null-space projection principle, a vector can be divided into two orthogonal parts: one in the null space of $\bm{A}_k \widetilde{\bm{P}}_k$ and the other in the column space of $\bm{A}_k \widetilde{\bm{P}}_k$. The component within the column space can be adjusted freely, while the part in the null space remains uncontrollable, as it reflects the influence of higher-priority constraints. Therefore, to find the solution that minimizes the angle error, we perform an orthogonal decomposition on both $\bm{A}_k \bm{\mu}_2$ and $\bm{b}_k$, and compute their angle based on the decomposition results. 

The influence of constraints with higher priority on the current task level is represented by $\bm{A}_k \widetilde{\bm{u}}_{k}$. Projecting $\bm{A}_k \widetilde{\bm{u}}_{k}$ into the null space of $\bm{A}_k \widetilde{\bm{P}}_k$ leads to
\begin{equation}
\label{epsilon}
    \bm{\epsilon} = \bm{A}_k \widetilde{\bm{u}}_{k} - \bm{A}_k (\bm{A}_k \widetilde{\bm{P}}_k)^{\dagger} \bm{A}_k \widetilde{\bm{u}}_{k},
\end{equation}
where $\bm{\epsilon}$ represents the unchangeable influence from higher-level equality constraints. Therefore, the task performance at the $k$th level can be expressed as
\begin{equation}
\label{Amu1}
    \bm{A}_k \bm{\mu}_2 = \bm{\epsilon} + \bm{\zeta},
\end{equation}
where $\bm{\zeta}$ is in the column space of $\bm{A}_k \widetilde{\bm{P}}_k$ and $\bm{\epsilon}^T \bm{\zeta} = 0$. 

Similarly, we can project $\bm{b}_k$ into the null and column spaces of \(\bm{A}_k \widetilde{\bm{P}}_k\) as
\begin{equation}
\label{gamma}
    \bm{\gamma} = \bm{b}_k - \bm{A}_k (\bm{A}_k \widetilde{\bm{P}}_k)^{\dagger}\bm{b}_k
\end{equation}
\begin{equation}
\label{varphi}
    \bm{\varphi} = \bm{A}_k (\bm{A}_k \widetilde{\bm{P}}_k)^{\dagger}\bm{b}_k,
\end{equation}
where $\bm{\gamma}$ and $\bm{\varphi}$ are in the null space and column space of $\bm{A}_k \widetilde{\bm{P}}_k$, respectively, $\bm{b}_k = \bm{\gamma} + \bm{\varphi}$, and $\bm{\gamma}^T  \bm{\varphi} = 0$. 

\begin{figure}[!htbp]
    \centering
    \includegraphics[width=0.4\textwidth]{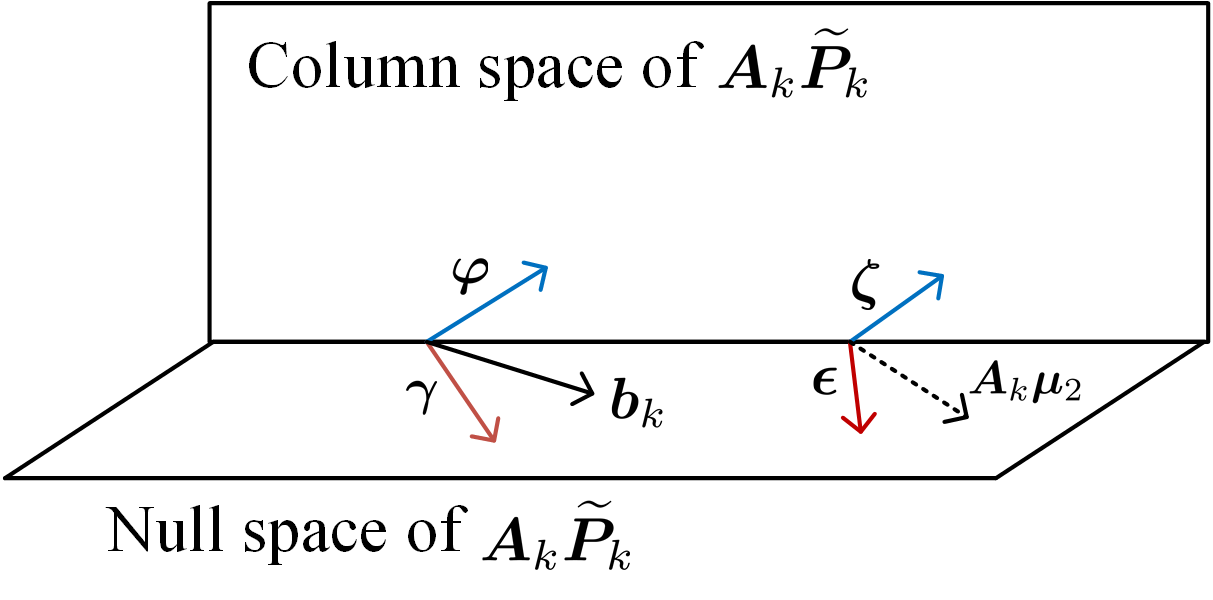}
    \caption{Projecting $\bm{b}_k$ and $\bm{A}_k \bm{\mu}_2$ into the null space and column space of $\bm{A}_k \widetilde{\bm{P}}_k$.}
    \label{projection}
\end{figure}

The equations \eqref{epsilon}, \eqref{gamma}, and \eqref{varphi} and the relationship between the projected vectors are illustrated in Fig. \ref{projection} for easy understanding.

Based on the projections, the problem of finding a solution with the minimum angle essentially becomes finding a $\bm{\zeta}$ following
\begin{equation}
    \mathop{\min}_{\bm{\zeta}} \alpha, ~ \alpha =  \angle((\bm{\gamma}+\bm{\varphi}),(\bm{\epsilon}+\bm{\zeta})).
\end{equation}
This problem is equivalent to maximizing the following $\cos \alpha$
\begin{equation}
\label{cosa}
\begin{aligned}
    \cos \alpha &= \frac{(\bm{\gamma}+\bm{\varphi}) (\bm{\epsilon}+\bm{\zeta})}{||\bm{b}_k|| \cdot||\bm{\epsilon}+\bm{\zeta}||}\\
    &=\frac{\bm{\gamma}^T \bm{\epsilon} + \bm{\varphi}^T \bm{\zeta}}{||\bm{b}_k|| \sqrt{\bm{\epsilon}^T \bm{\epsilon} + \bm{\zeta}^T \bm{\zeta}}}.
\end{aligned}
\end{equation}
According to (\ref{cosa}), when the norm of $\bm{\zeta}$ is fixed, the value of $\cos \alpha$ is determined only by $\bm{\varphi}^T \bm{\zeta}$. To maximize $\cos \alpha$, the direction of $\bm{\zeta}$ should be the same with that of $\bm{\varphi}$, as they are in the same space. Therefore, $\bm{\zeta}$ can be expressed as $s \bm{\varphi}$, where $s$ is a positive number. The objective then becomes finding a $s$ that maximizes
\begin{equation}
\label{fs}
    f(s) = \frac{\bm{\gamma}^T \bm{\epsilon} + s \bm{\varphi}^T\bm{\varphi}}{\sqrt{\bm{\epsilon}^T \bm{\epsilon} + s^2 \bm{\varphi}^T\bm{\varphi}}}.
\end{equation}
Differentiating (\ref{fs}) results in:
\begin{equation}
     f^\prime(s) = \frac{\bm{\varphi}^T\bm{\varphi} (\bm{\epsilon}^T \bm{\epsilon} - s \bm{\gamma}^T \bm{\epsilon})}{(\bm{\epsilon}^T \bm{\epsilon} + s^2 \bm{\varphi}^T\bm{\varphi})\sqrt{\bm{\epsilon}^T \bm{\epsilon} + s^2 \bm{\varphi}^T\bm{\varphi}}}.
\end{equation}
If $\bm{\gamma}$, $\bm{\epsilon}$ and $\bm{\varphi}$ are not zero, setting $f(s)^\prime = 0$ gives us:
\begin{equation}
    s = (\bm{\epsilon}^T \bm{\epsilon})/(\bm{\epsilon}^T \bm{\gamma}).
\end{equation}
If $\bm{\epsilon} = \bm 0$, $f(s) = 1$ always holds. If $\bm{\gamma} = \bm 0$, this means $\bm{b}_k$ can be achieved in the column space of $\bm{A}_k \widetilde{\bm{P}}_k$. In these two cases, we let $s = 1$. 

After obtaining $s$, we can replace $\bm{\zeta}$ with $s \bm{\varphi}$ in (\ref{Amu1}) as
\begin{equation}
    \bm{A}_k \bm{\mu}_2 = \bm{\epsilon} + s \bm{\varphi}.
\end{equation}
Substituting (\ref{epsilon}) and (\ref{varphi}) into the above equation, $\bm{\mu}_2$ can be computed as
\begin{equation}
\label{mu2}
    \bm{\mu}_2 = \widetilde{\bm{u}}_{k} + (\bm{A}_k \widetilde{\bm{P}}_k)^{\dagger}(s \bm{b}_k - \bm{A} \widetilde{\bm{u}}_{k}).
\end{equation}

Here, we need to ensure that \(\bm{\mu}_2\) does not violate any t-iq constraints. Recall that Algorithm \ref{al2} gives the possible scaling range for \(\bm{b}_k\). Therefore, \(s\) is checked to be within the feasible range \([s_{\min}, s_{\max}]\) and is adapted to the boundary if it is out of the range.

\emph{Remark 1}: The form of (\ref{mu2}) appears similar to the solution in \cite{eSNS}, which also features a scaling factor. However, the motivation behind our method is different. The scaling factor here is to find the solution with the minimum angle, while the method in \cite{eSNS} uses the factor to satisfy t-iq constraints. The proposed method can handle the cases when the rank of $\bm{A}_k \widetilde{\bm{P}}_k$ is less than the task dimension.

It should be noted that (\ref{mu2}) represents the optimal solution for minimizing $\angle (\bm{A}_k \bm{u}, \bm{b}_k)$, given the equality constraint set identified by the minimum error search. As the search progresses, the equality constraint set may change and lead to different solution to the minimum angle problem. To ensure local optimality, we compare solutions obtained from all historical equality constraint sets and select the most optimal one as the final outcome, as outlined in lines 22-26 of Algorithm \ref{al1}.

\subsection{Merging}
After obtaining \(\bm{\mu}_1\) and \(\bm{\mu}_2\), the final solution is computed as a linear combination of them as \(\eta\bm{\mu}_1 +  (1-\eta) \bm{\mu}_2\), where \(0 \leq \eta \leq 1\). The purpose of this process is to find a feasible solution under direction constraints. Since \(\bm{\mu}_1\) and \(\bm{\mu}_2\) satisfy all the t-iq constraints of levels 1 to \(k\), the linear combination of them also satisfies the t-iq constraints. In the merging process, the most important factor in determining the value of \(\eta\) is how the task direction and the task error vary with \(\eta\). We formulate the analysis as follows.

Let $\bm{\nu}_1 = \bm{A}_k \bm{\mu}_1$, $\bm{\nu}_2 = \bm{A}_k \bm{\mu}_2$ and  \(\beta = \angle(\bm{b}_k, \eta\bm{\nu}_1 +  (1-\eta) \bm{\nu}_2)\), we have:
\begin{equation}
\label{beta}
    \cos \beta = \frac{  \eta \bm{\nu}_1^T \bm{b}_k + (1-\eta) \bm{\nu}_2^T \bm{b}_k}{\sqrt{\eta^2 ||\bm{\nu}_1||^2 + 2\eta(1-\eta)\bm{\nu}_1^T\bm{\nu}_2 + (1-\eta)^2||\bm{\nu}_2||^2}||\bm{b}_k||}.
\end{equation}
To identify its changing trend with respect to \(\eta\), (\ref{beta}) can be simplified as the following function:
\begin{equation}
\label{gt}
    g(\tau) = \frac{\sigma_1 + \sigma_2 \tau}{\sqrt{\delta_2 \tau^2 + 2\rho \tau + \delta_1}},
\end{equation}
where \(\tau = (1-\eta)/\eta\), \(\rho = \bm{\nu}_1^T \bm{\nu}_2\), \(\delta_1 = \bm{\nu}_1^T \bm{\nu}_1\), \(\delta_2 = \bm{\nu}_2^T \bm{\nu}_2\), \(\sigma_1 = \bm{\nu}_1^T \bm{b}_k\), and \(\sigma_2 = \bm{\nu}_2^T \bm{b}_k\). The range of \(\tau\) is \([0,\infty]\) and it monotonically decreases with respect to \(\eta\).
Differentiating (\ref{gt}) yields:
\begin{equation}
\label{gd}
    g^\prime(\tau) =  \frac{(\rho \sigma_2 - \sigma_1 \delta_2)\tau + (\delta_1 \sigma_2 - \sigma_1 \rho)}{(\delta_2 \tau^2 + 2\rho \tau + \delta_1)\sqrt{\delta_2 \tau^2 + 2\rho \tau + \delta_1}}.
\end{equation}
Here, the sign of $g^\prime(\tau)$ is determined by its numerator, which is a linear function. It should be noted that \(\angle(\bm{b}_k, \bm{\nu}_1) \geq \angle(\bm{b}_k, \bm{\nu}_2)\) holds because $\bm{\mu}_2$ is obtained by the minimum angle solution search. This property implies $g(0) \leq g(\infty)$. Therefore, there are only two possible trends for \(g(\tau)\): monotonically increasing, or decreasing first and then increasing. In the first case, $\beta$ is monotonically increasing with respect to $\eta$. For the second case, we can compute a new $\bm{\mu}_2$ to convert it into the first case.
In the second case, \(\rho \sigma_2 - \sigma_1 \delta_2 < 0\) and \(\delta_1 \sigma_2 - \sigma_1 \rho \geq 0\). The minimum point can be computed as
\begin{equation}
\label{tau}
    \tau_m =  \frac{\sigma_1 \rho - \delta_1 \sigma_2}{\rho \sigma_2 - \sigma_1 \delta_2}.
\end{equation}
Then, we can obtain a new \(\bm{\mu}_2\) as:
\begin{equation}
\label{mu2_2}
    \bm{\mu}_2 \gets \frac{1}{1+\tau_m}\bm{\mu}_1 + \frac{\tau_m}{1+\tau_m}\bm{\mu}_2,
\end{equation}
and consequently a new $\bm{\nu}_2$ can be computed, where \(\angle(\bm{b}_k, \eta\bm{\nu}_1 +  (1-\eta) \bm{\nu}_2)\) is guaranteed to increase monotonically with respect to $\eta$.

On the other hand, let the task error be expressed as:
\begin{equation}
\begin{aligned}
        h(\eta) &= ||\eta \bm{\nu}_1 + (1-\eta)\bm{\nu}_2 - \bm{b}_k||\\
        &= ||\bm{\nu}_1||^2 \eta^2 + 2\eta(1-\eta)\bm{\nu}_1^T\bm{\nu}_2 + (1-\eta)^2||\bm{\nu}_2||^2\\
        & \quad - 2\eta \bm{\nu}_1^T \bm{b}_k - 2(1-\eta)\bm{\nu}_2^T \bm{b}_k + ||\bm{b}_k||^2.
\end{aligned}
\end{equation}
Using the same notations with (\ref{gt}), it can be simplified as follows:
\begin{equation}
\label{he}
    h(\eta) = (\delta_1 - 2\rho + \delta_2)\eta^2 + 2(\rho+\sigma_2 - \delta_2 - \sigma_1) \eta.
\end{equation}
Differentiating this equation leads to:
\begin{equation}
    h(\eta)^\prime = 2(\delta_1 - 2\rho + \delta_2)\eta + 2(\rho+\sigma_2 - \delta_2 - \sigma_1).
\end{equation}
The first derivative of \(h(\eta)\) is also a linear function, and \(\delta_1 - 2\rho + \delta_2 \geq 0\) always holds as it is equivalent to \(||\bm{\nu}_1 -  \bm{\nu}_2||^2\). As \(\bm{\mu}_1\) is the solution minimizing the task error globally, \(h(1)\) is the minimum value when \(\eta \in [0, 1]\), and implies \(h(\eta)\) must be monotonically decreasing in \([0, 1]\).

Combining the above two aspects, as \(\eta\) grows, the solution \(\eta\bm{\mu}_1 +  (1-\eta) \bm{\mu}_2\) can achieve a smaller task error while the task direction deviation will be larger. Therefore, the best $\eta$ is achieved at the boundary of the direction constraint, which satisfies
\begin{equation}
    \angle(\eta \bm{\nu}_1 + (1-\eta) \bm{\nu}_2, \bm{b}_k)=\theta_k.
\end{equation}
The pseudocode of the solution merging process is presented in Algorithm \ref{al3}.

\begin{algorithm}[!htb]
    \caption{Solution merging process}
    \label{al3}
    \begin{algorithmic}[1]
    \Function{merge}{$\bm{\mu}_1,\bm{\mu}_2, \theta_k$}
    \State $\bm{\nu}_1 = \bm{A}_k \bm{\mu}_1, \bm{\nu}_2 = \bm{A}_k \bm{\mu}_2$
    \State $\rho = \bm{\nu}_1^T \bm{\nu}_2$
    \State $\delta_1 = \bm{\nu}_1^T \bm{\nu}_1, \delta_2 = \bm{\nu}_2^T \bm{\nu}_2$
    \State $\sigma_1 = \bm{\nu}_1^T \bm{b}_k, \sigma_2 = \bm{\nu}_2^T \bm{b}_k$
    \State $\tau =  (\sigma_1 \rho - \delta_1 \sigma_2)/(\rho \sigma_2 - \sigma_1 \delta_2) $
    \If{$\tau > 0$ \textbf{and} $\rho \sigma_2 - \sigma_1 \delta_2 < 0$}
        \State$\bm{\mu}_2 \gets (\frac{1}{1+\tau})\bm{\mu}_1 + (\frac{\tau}{1+\tau})\bm{\mu}_2$
        \State $\bm{\nu}_2 = \bm{A}_k \bm{\mu}_2$
    \EndIf
    \If{$\angle(\bm{\nu}_1,\bm{b}_k) \leq \theta_k$}
        \State \Return $\bm{\mu}_1$
    \ElsIf{$\angle(\bm{\nu_}2,\bm{b}_k) \geq \theta_k$}
        \State \Return $\bm{\mu}_2$
    \Else
        \State Compute $\eta$ using $\angle(\bm{A}_k(\eta \bm{\nu}_1 + (1-\eta) \bm{\nu}_2), \bm{b}_k)=\theta_k$ 
        \State \Return $\eta \bm{\mu}_1 + (1-\eta) \bm{\mu}_2$
    \EndIf
    \EndFunction
    \end{algorithmic}
\end{algorithm}

\section{Using Variable Admittance to Improve Interaction Efficiency at Constraint Boundaries} \label{Variable}

In the previous section, we assumed that the desired end-effector velocity is derived from a general admittance controller, as represented in \eqref{ad2}. However, the velocity output \(\bm{v}_a\) in \eqref{ad2} is solely influenced by the interaction force. An increase in the force applied by the human operator leads to an increase in \(\bm{v}_a\), even when the robot velocity is constrained (i.e., certain t-iq constraints become active). Such conflict between admittance and constraints may significantly reduce human-robot interaction efficiency. For instance, if the human operator continuously increases the force and pushes the robot to the boundary of a position constraint, its velocity will become zero due to t-iq constraints while \(\bm{v}_a\) continues to rise. Then, if the human changes the force to the opposite direction, it will take some time for \(\bm{v}_a\) to decrease to zero before increasing in the new direction. The robot will not immediately follow the human force although its velocity is zero. This boundary problem has been rarely discussed in previous literature on admittance control, as most studies assume that the velocity command of the admittance control can be perfectly achieved. In this section, we present a variable admittance strategy to address this problem.

Essentially, we use a variable damping coefficient in addition to a constant mass to achieve better stability and performance \cite{2f}. Based on (\ref{ad2}), the variable admittance model can be written as
\begin{equation}
    \label{ad3}
    \bm{M} \dot{\bm{v}}_a + \bm{D}(t) \bm{v}_a = \bm{f}_{ext}.
\end{equation}
Let $\bm{\chi} = \bm{v}_a - \bm{v}$ denote the deviation between the velocity objective obtained through the admittance controller and the velocity of the robot's end-effector. Our proposed method is to adjust the damping coefficient by considering this velocity deviation alongside the human intention. The update law of the damping term \(\bm{D}(t)\) is designed as
\begin{equation}
\label{vary}
    D_i = \mathrm{max}\{ \kappa_1 \mathrm{e}^{\mathrm{sign}(f_{ext,i}) \kappa_2 \chi_i}, D_{min} \},
\end{equation}
where \(D_i\) is the \(i\)th element on the diagonal of \(\bm{D}(t)\), \(\kappa_1\) and \(\kappa_2\) are two positive gains, \(D_{min}\) is a lower bound, and \(\mathrm{sign}(f_{ext,i})\) denotes the sign of the \(i\)th element of the external force.

\begin{figure}
    \centering
    \includegraphics{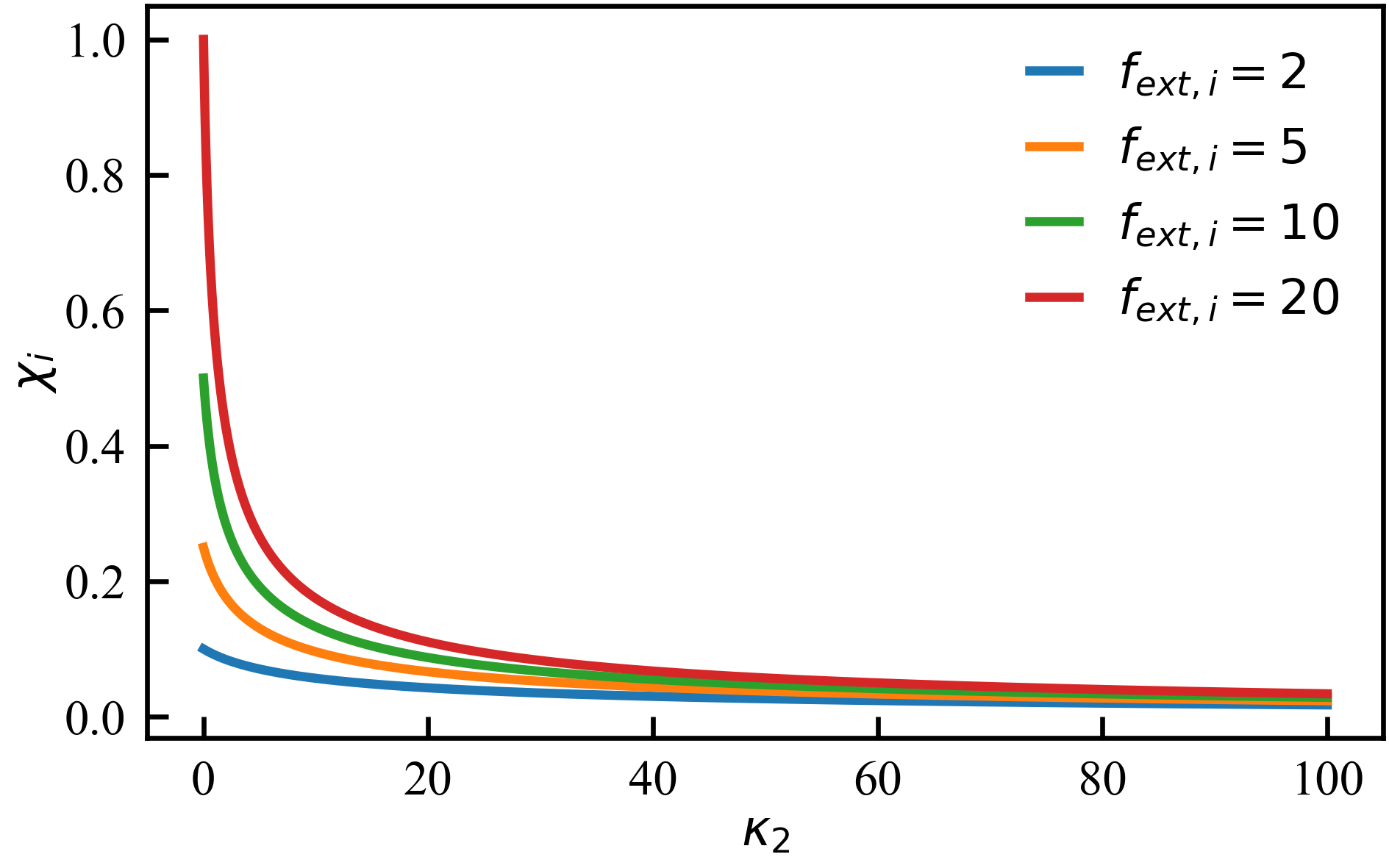}
    \caption{Relation between \(\chi_i\) and \(\kappa_2\) under the conditions \(v_i=0\), \(\kappa_1=20\), \(f_{ext,i}=2,5,10,20\)}
    \label{chi}
\end{figure}

When \(\bm{\chi} = 0\), the velocity \(\bm{v}_a\) can be achieved under the t-iq constraints. In this case, \(\bm{D}(t)\) is constant and determined by the value of \(\kappa_1\). However, when \(\bm{\chi} \neq 0\), some constraints are active and saturated at their boundaries. If \(\mathrm{sign}(f_{ext,i}) \chi_i > 0\), the human is driving the end-effector in the direction violating the constraints. Hence, the damping is increased to prevent \(\bm{v}_a\) from growing. Conversely, if \(\mathrm{sign}(f_{ext,i}) \chi_i < 0\), the human is driving the end-effector away from violating the constraints. Then, the damping is decreased to allow \(\bm{v}_a\) to change rapidly.

Further analysis of the effect of (\ref{vary}) can be performed under the condition of a constant external force. According to (\ref{ad3}), the equilibrium point of the admittance controller is achieved when
\begin{equation}
      v_{a,i} = \frac{f_{ext,i}}{D_i}.
\end{equation}
By substituting (\ref{vary}) into this equation, we obtain
\begin{equation}
    (\chi_i + v_i) \kappa_1 \mathrm{e}^{\mathrm{sign}(f_{ext,i}) \kappa_2 \chi_i} = f_{ext,i}.
\end{equation}
Although the equation does not have an analytical solution for \(\chi_i\), it can be inferred that \(\chi_i\) approaches zero when \(\kappa_2\) is appropriately selected. Fig. \ref{chi} illustrates the relationship between \(\chi_i\) and \(\kappa_2\) under the conditions \(v_i=0\), \(\kappa_1=20\), and \(f_{ext,i}=2,5,10,20\) as an example. The figure shows that the velocity deviation decreases with increasing $\kappa_2$, and across all cases, the deviation remains small and near zero when $\kappa_2$ is appropriately chosen (e.g., $\kappa_2 = 60$). The analysis demonstrates that the proposed variable damping is able to handle the velocity deviation problem caused by t-iq constraints. The components of the desired velocity in the constrained direction are suppressed. It can improve interaction performance when the human operator intends to move along the constraint boundary. The direction of \(\bm{v}_a\) becomes closer to the direction of unrestricted motion, making it easier to satisfy the direction constraints. The effectiveness of this improvement will be examined in more detail in the experimental section.
\section{Experiments and Analysis} \label{Experiment}
In this section, four case studies are conducted to validate the proposed method. In the first two case studies, we compare the pHRI performance of our method with that of the HQP and task scaling methods. In the third case study, we evaluate the robot performance with and without variable admittance. Finally, we present a case study where a robot assists a human in assembling two pieces of aluminum profiles. The experiments were performed on the UFACTORY xArm7, a 7-degree-of-freedom robotic arm. The computational device used in our experiments was a PC with an Intel Core i7 CPU and 16 GB of memory. Detailed experimental setup and comparisons can be viewed in the supplementary video. 

\begin{figure}[!htbp]
    \centering
    \includegraphics[width=0.45\textwidth]{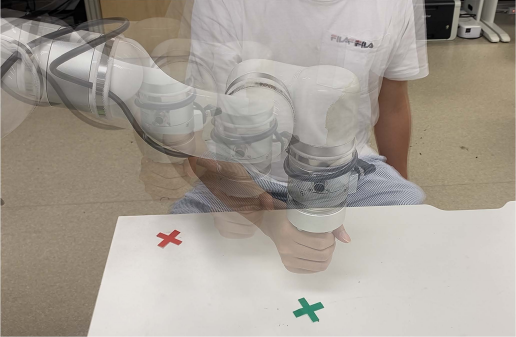}
    \caption{First case study. The red cross is the start point and the green one is the desired point.}
    \label{exp_img}
\end{figure}

\begin{figure}[!htbp]
    \centering
    \includegraphics[width=0.45\textwidth]{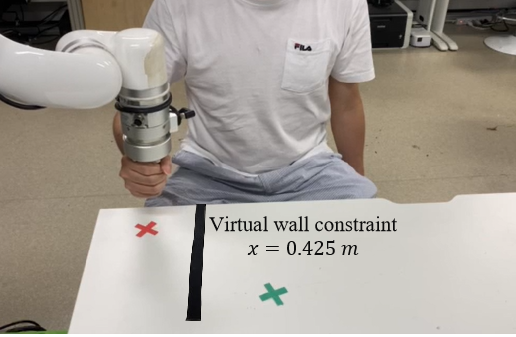}
    \caption{Second case study. The red cross is the start point and the green one is the desired point. The black line is the virtual wall constraint at $x = 0.425 ~ m$.}
    \label{exp_img2}
\end{figure}


\subsection{First Case Study for Comparing Different Methods}
\begin{figure*}
    \centering
    \includegraphics[width=0.85\textwidth]{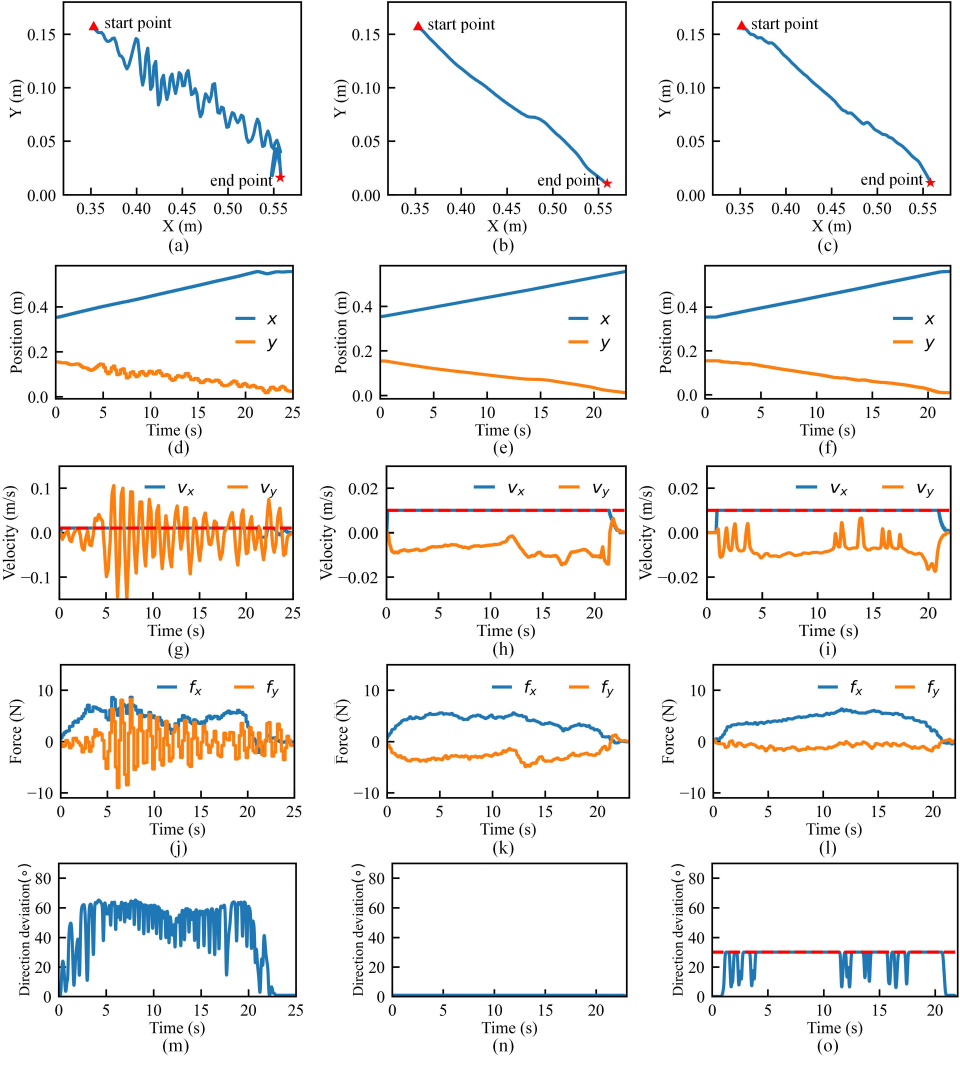}
    \caption{Results of the first case study: Left column -- HQP; Middle column -- Task scaling; Right column -- Proposed direction-constrained optimization. (a)-(c) Trajectory of the end-effector in the x-y plane. (d)-(f) Time history of the end-effector's position. (g)-(i) Time history of the end-effector's velocity. (j)-(l) Time history of the interaction force. (m)-(o) Time history of the direction deviation during the human-robot interaction subtask.}
    \label{cmp1}
\end{figure*}

\begin{figure*}
    \centering
    \includegraphics[width=0.85\textwidth]{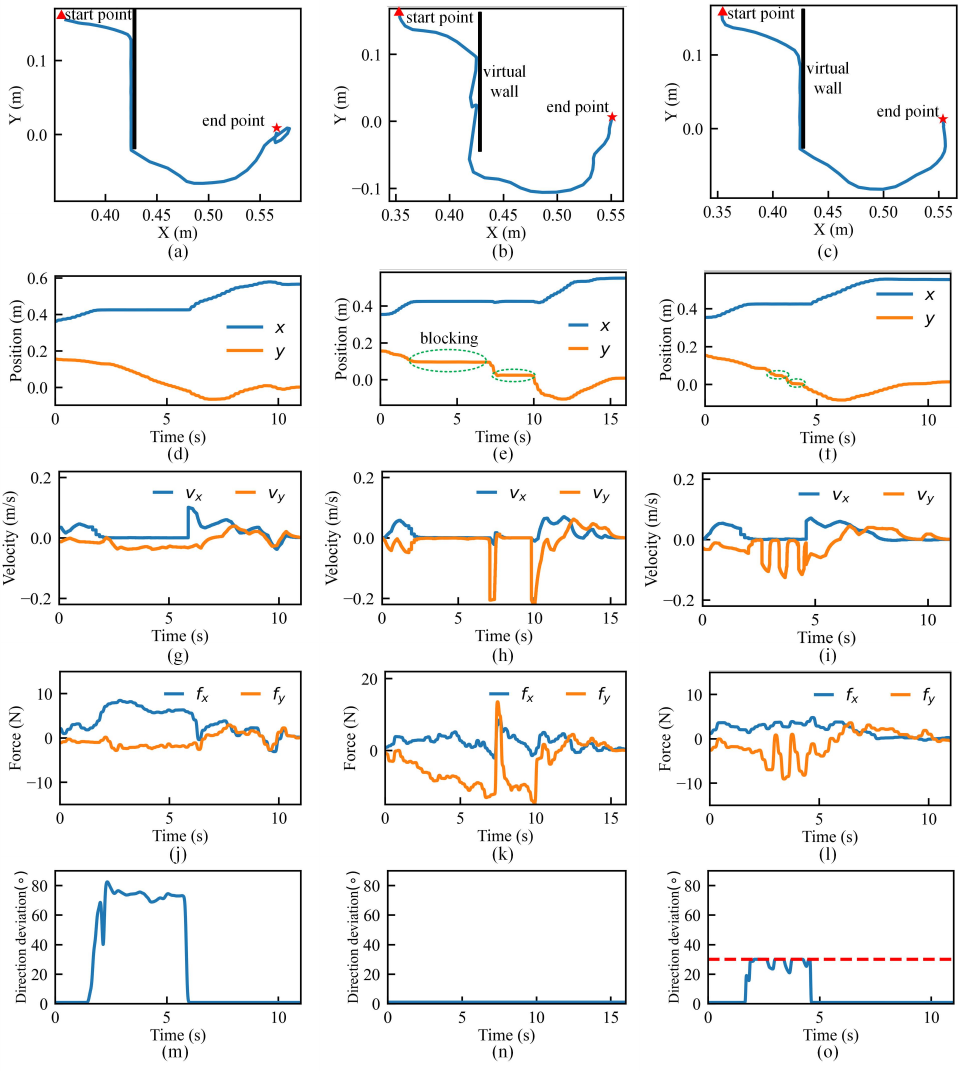}
    \caption{Results of the second case study: Left column -- HQP; Middle column -- Task scaling; Right column -- Proposed direction-constrained optimization. (a)-(c) Trajectory of the end-effector in the x-y plane. (d)-(f) Time history of the end-effector's position. (g)-(i) Time history of end-effector's velocity. (j)-(l) Time history of the interaction force. (m)-(o) Time history of the direction deviation during the human-robot interaction subtask.}
    \label{cmp2}
\end{figure*}

In the first case study, we compare the performance of the HQP \eqref{op1}, task scaling \eqref{op-scale} and direction-constrained optimization (\ref{op2}) in a pHRI task with velocity constraints for the end-effector.  The task comprises two levels of priority. The first level involves maintaining the initial orientation and height of the end-effector, with joint position limits as t-iq constraints. The equality constraint of the second level involves a two-dimensional reference velocity (in the x-y plane) derived from an admittance controller with constant parameters. The t-iq constraint for this level is to keep the end-effector's velocity along the x-axis below 0.01 m/s. Additionally, in the direction-constrained optimization, the task angle deviation for the interaction task is limited to within $30^\circ$, while no direction constraint is applied at the first level.  In the experiments, a human physically interacts with the robot's end-effector to move it from a start point to a desired point. Snapshots of the interaction process are shown in Fig. \ref{exp_img}.

The experimental results are presented in Fig. \ref{cmp1}. During the interaction, as the human pushes the robot in the desired direction, the reference velocity from the admittance controller increases. However, for most of the process, the velocity constraint along the x-axis is activated and saturated at 0.01 m/s. When using the HQP method, the velocity along the y-axis is unconstrained and can be much larger than 0.01m/s, causing the robot's moving direction to deviate significantly from the human's intention. As shown in Fig. \ref{cmp1}(a), the trajectory keeps oscillating during interaction. The angles between the end-effctor's velocity and the velocity objective from the admittance controller approach $60^\circ$. It significantly reduces human comfort as considerable effort must be exerted to adjust the force frequently. This explains the intensive oscillations in the velocity and interaction force of Fig. \ref{cmp1}(g) and (j). 

In contrast, the robot's motion is more stable using the task-scaling and direction-constrained optimization. By constraining the direction of the robot's motion, the velocity constraint along the x-axis also limits the velocity increase along the y-axis. As a result, the robot's trajectory is smoother, and human comfort is improved. For the direction-constrained optimization, the velocity direction deviation is successfully limited within $30^\circ$. There are no significant differences between these two methods in this case study. These results demonstrate that considering robot's moving direction in pHRI is crucial, and restricting the direction with a certain range is feasible.

\subsection{Second Case Study for Comparing Different Methods}
This set of experiments compares the three methods in a pHRI task with position constraints. The interaction task for the human is also to drive the robot's end-effector to a desired point in a plane. Additionally, a virtual wall obstacle, positioned at $x=0.425$ m, $y\geq -0.02$ m between the two points, is introduced as a position constraint, as shown in Fig. \ref{exp_img2}. However, the human participant was unaware of the exact location of this virtual constraint. Consequently, during the task, the human initially moved toward the virtual wall and then along it, attempting to navigate around the obstacle as quickly as possible.

The task design includes two levels of priority, similar to the previous experiment. However, in this case, we add the virtual wall constraint as a t-iq constraint at the first level and remove the velocity constraint from the second level. The task angle deviation for the interaction task is also limited to within $30^\circ$ in the direction-constrained optimization.

Fig. \ref{cmp2} presents the experimental results of the three methods. With the task scaling method, it is observed that while the task direction is preserved, the human found it difficult to move along the boundary of the virtual wall. The trajectory results show that the robot could not move closely along the virtual wall, necessitating a backward motion, as shown in Fig. \ref{cmp2}(a). Additionally, the robot motion was blocked for an extended period, approximately 5 seconds (circled in green in Fig. \ref{cmp2}(e)), upon reaching the virtual constraint boundary. During this time, the human had to continuously adjust his force, and the interaction efficiency is reduced. This issue arises because, at the boundary, the velocity towards the virtual wall is constrained to nearly zero, causing the scaling factor to be zero and consequently scaling all components of the command velocity to zero.

However, in this case study, the HQP outperforms the task scaling. This is because the HQP method only needs to preserve the velocity components along the virtual wall, allowing the velocity direction to be arbitrary. In addition, our method also addresses the problem of the task scaling by accepting velocities within the predefined direction range. As shown in Fig. \ref{cmp2}(c), the robot can move smoothly along the virtual wall. Although the robot motion is occasionally blocked, the human can easily adapt their force to continue moving, and the blocking duration is significantly shorter than with the task scaling method. The blocking problem at the boundary can be further improved using variable admittance, as will be demonstrated in the next subsection.

These two case studies demonstrate the capability of the proposed method to handle different scenarios. In the first case study, considering the robot's moving direction helps improve user experience. However, as shown in the second case study, strictly limiting the direction can reduce efficiency. While the HQP and task scaling methods perform well in only one of the cases, our method is effective in both. 

\subsection{Variable Admittance vs. Constant Admittance}
This set of experiments demonstrates the advantage of using variable admittance in the direction-constrained optimization. Two groups of comparative experiments were conducted, both with two levels of priority. The setting is almost the same as in the previous subsection, except that the virtual wall extends infinitely along \(x = 0.425\) $m$.

\begin{figure}[!htbp]
    \centering
    \includegraphics[width=0.48\textwidth]{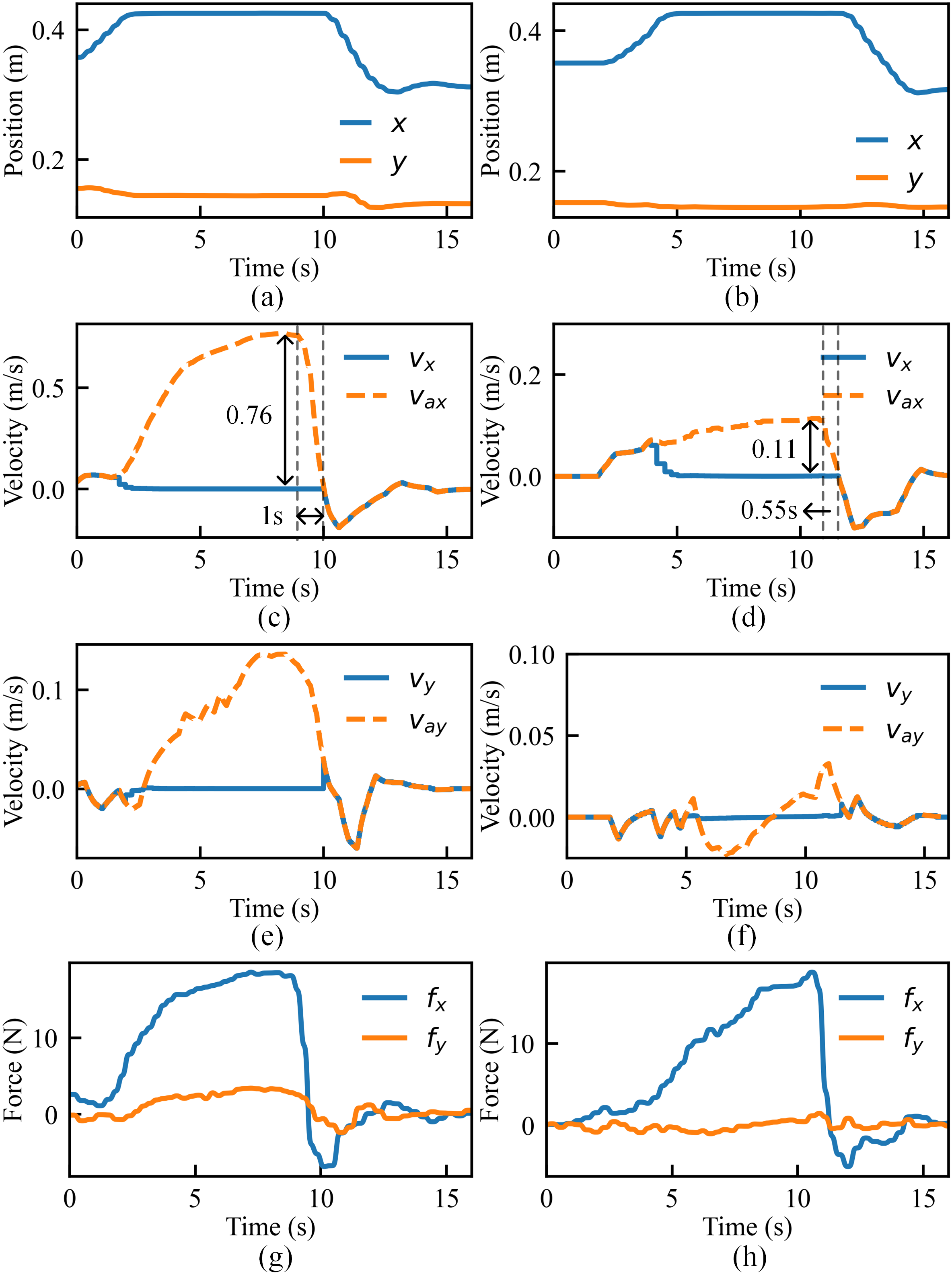}
    \caption{The first group of comparison results between fixed admittance (left column) and variable admittance (right column). (a)-(b) Time history of the end-effector's position. (c)-(d) Time history of the velocity objective from the admittance controller and the end-effector's velocity along x-axis. (e)-(f) Time history of the velocity objective from the admittance controller and the end-effector's velocity along y-axis. (g)-(h) Time history of the interaction force. }
    \label{cmp3}
\end{figure}

In the first experiment, the human pushed the end-effector towards the virtual wall and then moved in the opposite direction. The parameters for the fixed admittance controller were chosen as $\bm{M}=10\bm{I}$ and $\bm{D}=30\bm{I}$. For the variable admittance controller, the parameters were $\bm{M}=10\bm{I}$, $\kappa_1 = 30$, $\kappa_2 = 30, D_{min} = 20$. The results are shown in Fig. \ref{cmp3}. With the fixed damping, the velocity $\bm{v}_a$ from the admittance controller rapidly increased as the human force grew. However, the end-effector's velocity was constrained to zero, resulting in a large velocity deviation along the x-axis of 0.76 m/s. This deviation was only 0.11 m/s when using variable damping. As shown in Figs. \ref{cmp3}(c)-(d), when the human exerted force in the opposite direction away from the constraint, it took about 1 second to follow the human intention with fixed damping, whereas it took only 0.55 seconds with variable damping. These results validate that the proposed variable admittance controller can improve interaction efficiency when constraints are active.

\begin{figure}[!htbp]
    \centering
    \includegraphics[width=0.48\textwidth]{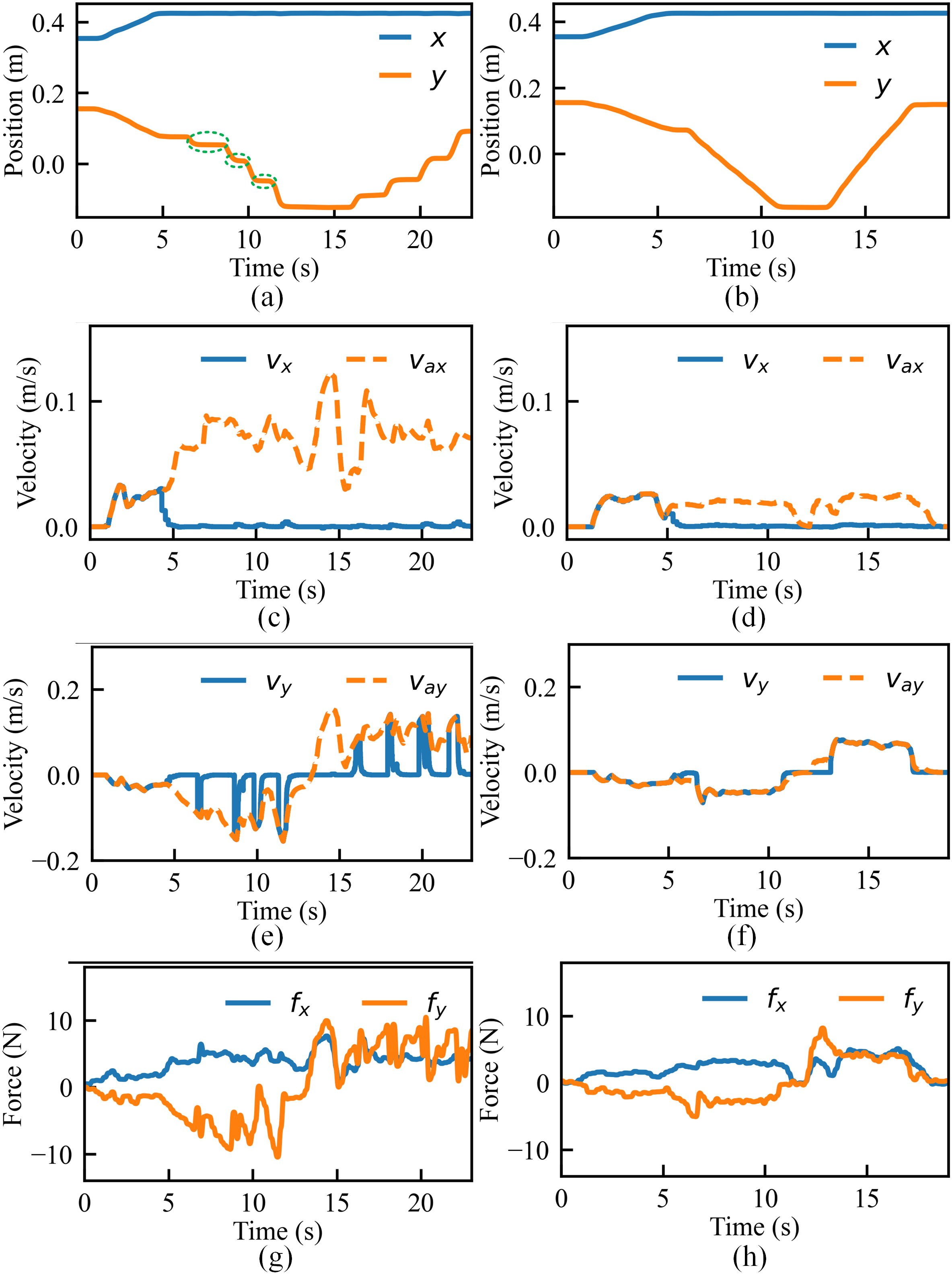}
    \caption{The second group of comparison results between fixed admittance (left column) and variable admittance (right column). (a)-(b) Time history of the end-effector's position. (c)-(d) Time history of the velocity objective from the admittance controller and the end-effector's velocity along x-axis. (e)-(f) Time history of the velocity objective from the admittance controller and the end-effector's velocity along y-axis. (g)-(h) Time history of the interaction force. }
    \label{cmp4}
\end{figure}

In the second experiment, the human attempted to drive the robot back and forth along the virtual wall constraint at its boundary. The results are presented in Fig. \ref{cmp4}. As shown in Figs. \ref{cmp4}(a)-(b), the robot motion was blocked several times under fixed damping, while smooth movement was achieved with variable damping. The blocking problem occurs because no feasible solution can be found under the direction constraint. As the variable admittance controller accounts for the constraints, its output is regulated by the deviation between the robot velocity and command velocity, as shown in Figs. \ref{cmp4}(c)-(f). This adjustment makes the velocity direction from the variable admittance controller closer to the unconstrained direction, thereby reducing the effort required by the human to interact with the robot at the constraint boundary.

\subsection{Human–Robot Co-Assembly under Visual Guidance}
\begin{figure}[!htbp]
    \centering
    \includegraphics[width=0.44\textwidth]{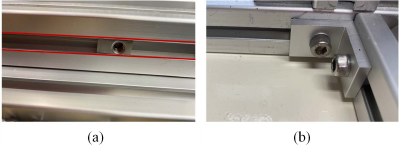}
    \caption{Experimental settings of the co-assembly task. (a) A single aluminum profile with a hole in the middle frame slot. The red lines are line features used in the experiment. (b) Two aluminum profiles connected together.}
    \label{profile}
\end{figure}

This subsection presents an application of the proposed method in human-robot cooperative assembly. Specifically, we consider a scenario in which a robot assists a human in assembling two aluminum profiles. A target hole is placed in the middle frame slot of the aluminum profile, as shown in Fig. \ref{profile}(a). Usually, there can be several holes in the frame slot, and the human decides which one to use. In this task, line features of the profile are utilized as partial knowledge with the visual servoing technique \cite{vs}. These lines are extracted by an in-hand camera. It is assumed that the center of the aluminum profile attached to the robot is aligned with the camera center.

The cooperation task is designed with three levels of priority. The first subtask controls the z-axis of the end-effector to align with the normal vector of the target profile plane, with joint position limits as t-iq constraints. The second level is defined in the image space. Image features are computed from the two parallel lines of the frame slot. The first image feature is the orientation of the lines, and the second is the sum of the distances from the center to the two lines in the normalized image plane. For this task, the line orientation is controlled to be zero, and the second image feature is a t-iq constraint within the range of $[-0.1,0.1]$. Details on computing the Jacobian matrix of the image features can be found in \cite{line}. With these first two levels, the robot can roughly align the two profiles. The final subtask is to achieve the three-dimensional velocity of the variable admittance controller for compliant interaction with the human.  The velocity of the end-effector is limited to 0.03m/s by a t-iq constraint at this level. A varying direction constraint is designed for this task as follows:
\begin{equation}
\label{theta3}
    \theta_3(t)= max(\frac{d-d_e}{d_s-d_e} 45^\circ, 10^\circ)
\end{equation}
where $d$ is the distance between the camera and the target profile, $d_s$ is their distance at the beginning and $d_e$ is an estimate of their distance at the end. The idea of this design is to limit the direction of the robot motion more when precise operation is required. Table \ref{table1} summarizes the task hierarchy of the experiment.

\begin{table}[!htbp]
\footnotesize
    \centering
    \caption{Summary of the Task Hierarchy in the Human-Robot Co-Assembly Experiment}
    \renewcommand{\arraystretch}{1.5}
 \begin{tabular}{m{0.04\textwidth}<{\centering}|m{0.14\textwidth}<{\centering}|m{0.14\textwidth}<{\centering}|m{0.07\textwidth}<{\centering}}
\hline
Priority level & Subtasks & t-iq constraints & Direction constraint\\
\hline
1 & Orientation of z-axis of the end-effector & Joint position limits & $\theta_1 =  \pi$\\
\hline
2 & Orientation of the line feature & Distances from the center to two lines in the image & $\theta_2 =  \pi$\\
\hline
3 & Velocity from the admittance controller & Velocity of the end-effector & Defined by (\ref{theta3}) \\
\hline
\end{tabular}
    \label{table1}
\end{table}

\begin{figure}[!htbp]
    \centering
    \includegraphics[width=0.48\textwidth]{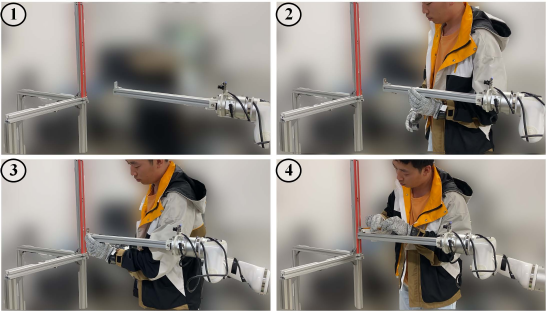}
    \caption{Snapshots of the human–robot co-assembly experiment. The numbers in the top-left corner of each picture indicate specific moments as follows: \ding{172} initial pose of the robot before interaction, \ding{173} human begins to guide the robot, \ding{174} human aligns the holes on the profiles, \ding{175} human fastens the two profiles using a screw.}
    \label{assemble_img}
\end{figure}

\begin{figure}[!htbp]
    \centering
    \includegraphics[width=0.48\textwidth]{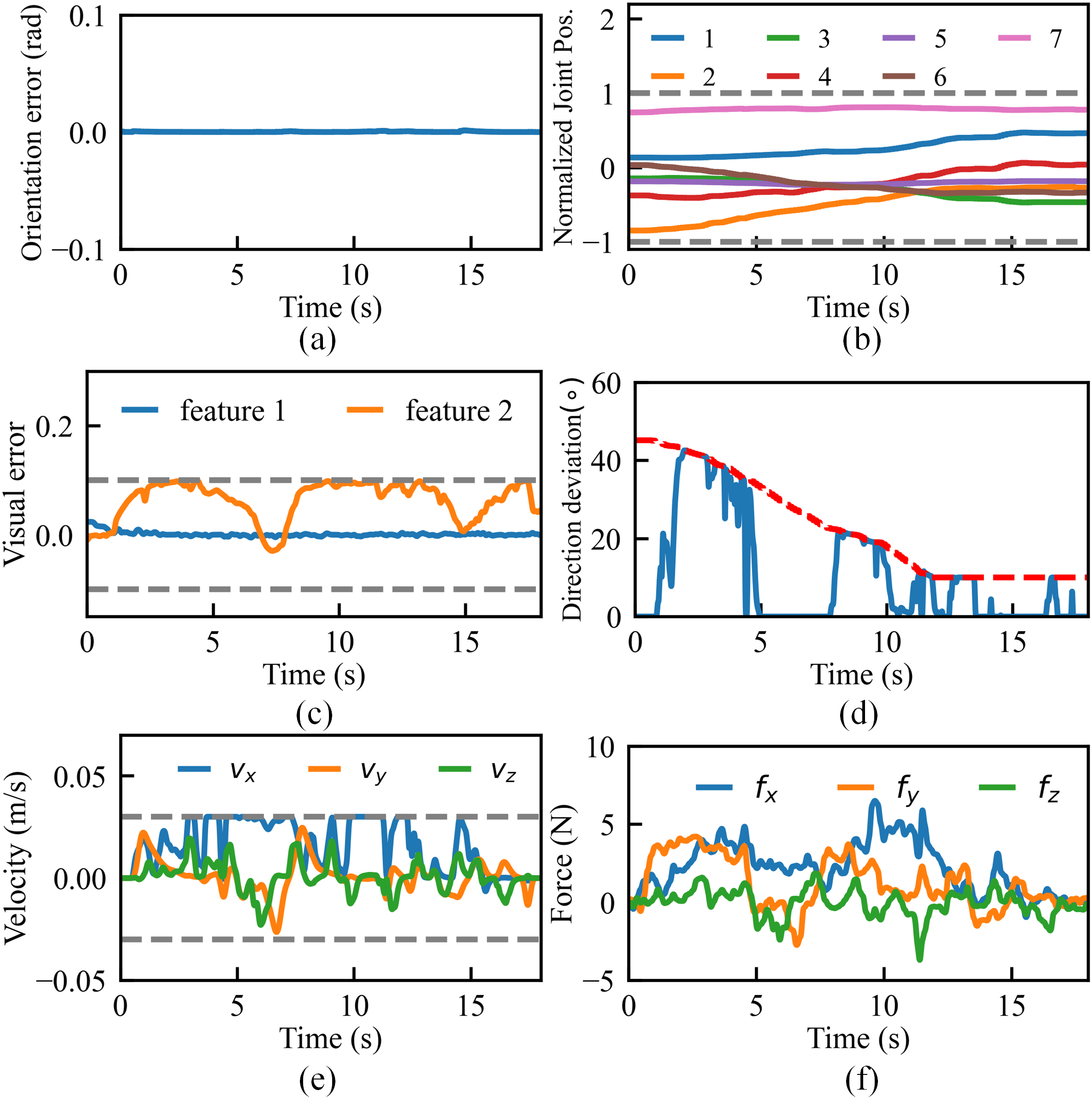}
    \caption{Experimental results of the human–robot co-assembly experiment. (a) Time history of the angle error of the z-axis of the end-effector. (b) Normalized joint position to the interval [-1, 1].  (c) Time history of image feature errors. (d) Time history of the direction deviation of the human-robot interaction subtask. (e) Time history of the robot velocity. (f) Time history of the interaction force. }
    \label{exp4}
\end{figure}

The experiment process is demonstrated in Fig. \ref{assemble_img}, and the results are reported in Fig. \ref{exp4}. During the cooperation, the subtasks and t-iq constraints of the first two levels are satisfied. The direction constraint of the third level remains within the predefined range. These results validate the effectiveness of our method in handling a complex pHRI task.

\section{Conclusions} \label{Conclusion}
In this paper, we addressed the problem of physical human-robot interaction (pHRI) under hierarchical constraints, with particular attention paid to boundary constraints. We developed a hierarchical control framework by integrating direction-constrained optimization and a variable admittance controller to enhance the performance of the robot.

In the experiments, we first compared our method with existing methods, including HQP and task scaling methods. Compared to HQP, our method showed stable robot behavior, whereas HQP exhibited oscillations. Compared to task scaling methods, our method allowed the robot to easily move along the constraint boundary in pHRI. Secondly, we compared variable admittance with fixed admittance. The results demonstrated that variable admittance achieves smoother motion along the constraint boundary. In addition, we conducted a human-robot co-assembly task with three levels of priority and demonstrated the effectiveness of our proposed method in the context of a practical application.

When all hard constraints are met, the performance of our proposed method is comparable to that of the HQP method. The benefits of our approach become more apparent when the robot reaches the boundaries of the hard constraints. At the boundaries, the direction constraints are activated, which restrict the robot’s movement direction and prevent significant and unpredictable deviations. Our proposed method improves both interaction comfort and cooperation efficiency from this viewpoint.


\bibliographystyle{IEEEtran}
\bibliography{IEEEabrv,reference}

\end{document}